\documentclass{article}
\usepackage[T1]{fontenc}
\usepackage[utf8]{inputenc}
\PassOptionsToPackage{table}{xcolor}
\usepackage{main}
\usepackage{microtype}
\usepackage{subcaption}
\usepackage{graphicx}
\usepackage{newtxtext}  
\usepackage{amsmath}
\usepackage{amssymb}
\usepackage{amsfonts}
\usepackage{newtxmath}  
\usepackage{float}
\usepackage{footnote}
\usepackage{enumitem}
\usepackage{bm}
\usepackage{booktabs}
\usepackage{array}
\usepackage{tabularx}
\usepackage{multicol}
\usepackage{multirow}
\usepackage{colortbl}
\usepackage{pifont}
\usepackage{makecell}
\usepackage{mathtools}
\usepackage{longtable}
\usepackage{rotating}
\usepackage{arydshln}
\usepackage{natbib}
\usepackage[normalem]{ulem}
\definecolor{mydarkblue}{rgb}{0,0.08,0.45}
\usepackage[colorlinks=true,linkcolor=mydarkblue,citecolor=mydarkblue,filecolor=mydarkblue,urlcolor=mydarkblue]{hyperref}
\usepackage{CJKutf8}
\usepackage{geometry}
\usepackage{pdflscape}
\usepackage{wrapfig}
\usepackage{pgfplots}
\usepackage{pgfplotstable}
\pgfplotsset{compat=1.18}

\usepackage{fancyhdr}

\fancypagestyle{titlepage}{%
  \fancyhf{}%
  \lhead{\raisebox{-0.3cm}{\includegraphics[height=0.95cm]{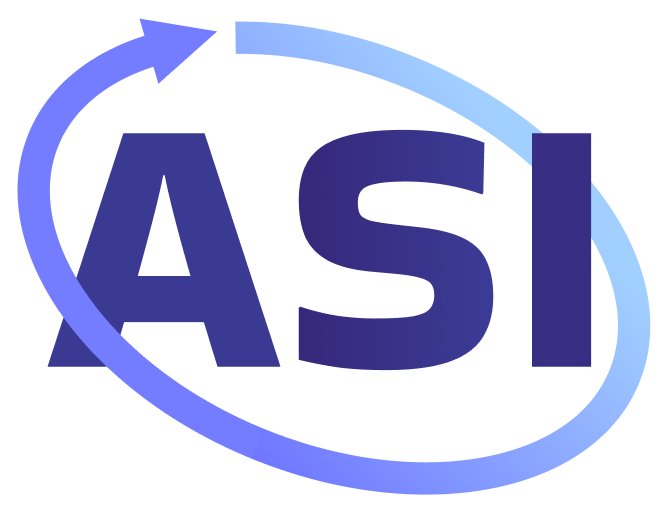}}}%
  \rhead{\raisebox{-0.2cm}{\includegraphics[height=0.7cm]{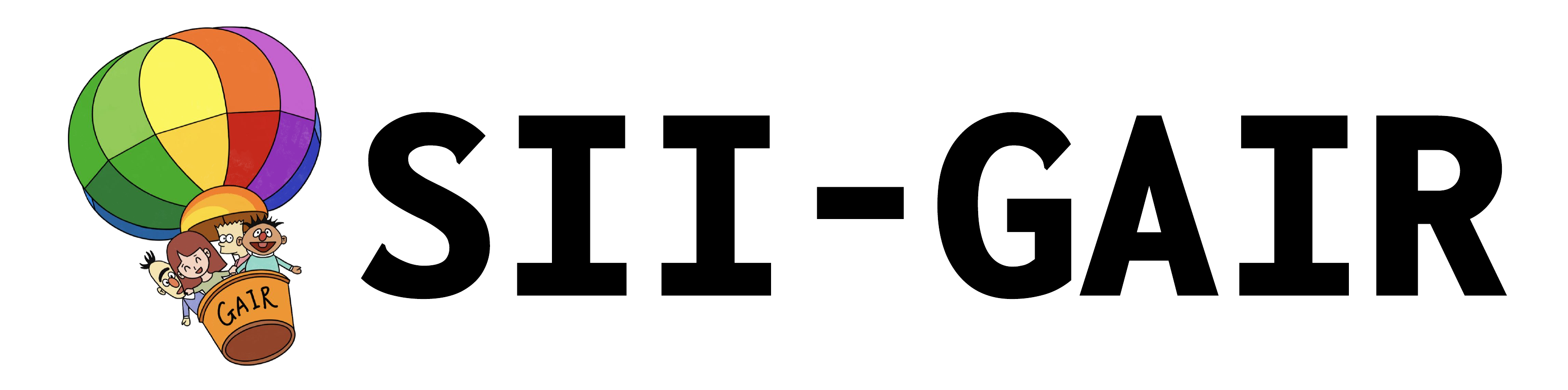}}}%
  \cfoot{\thepage}%
}

\fancypagestyle{normal}{%
  \fancyhf{}%
  \lhead{\rightmark}%
  \cfoot{\thepage}%
}

\DeclareCaptionFont{black}{\color{black}}
\newcommand{\cmark}{\ding{51}}
\newcommand{\xmark}{\ding{55}}

\definecolor{tablegreen}{rgb}{0.82, 0.94, 0.75}
\definecolor{tableblue}{rgb}{0.81, 0.90, 0.94}
\definecolor{tablered}{rgb}{0.97, 0.85, 0.85}
\definecolor{tableorange}{rgb}{0.96, 0.85, 0.81}

\newenvironment{itemize*}%
 {\leftmargini=10pt\begin{itemize}%
  \setlength{\itemsep}{0pt}%
  \setlength{\parskip}{0pt}%
  }%
 {\end{itemize}}
\newenvironment{enumerate*}%
 {\begin{enumerate}%
  \setlength{\itemsep}{0pt}%
  \setlength{\parskip}{0pt}}%
 {\end{enumerate}}

\usepackage{etoolbox}
\newcounter{bibcount}
\makeatletter
\patchcmd{\@lbibitem}{\item[}{\item[\hfil\stepcounter{bibcount}{[\thebibcount]}}{}{}
\renewcommand\NAT@bibsetup%
  [1]{\setlength{\leftmargin}{\bibhang}\setlength{\itemindent}{-\parindent}%
      \setlength{\itemsep}{\bibsep}\setlength{\parsep}{\z@}}
\makeatother

\makeatletter
\renewcommand{\@toptitlebar}{%
  {\color{black}\hrule height 1\p@}%
  \vskip 0.05in
  \vskip -\parskip%
}
\renewcommand{\@maketitle}{%
  \vbox{%
    \hsize\textwidth
    \linewidth\hsize
    \vskip 0.25in  
    \@toptitlebar
    \vskip 0.15in
    {\centering\LARGE\bfseries \@title\par}%
    \vskip 0.15in
    {\raggedright\@author\par}%
    \vskip 0.3in \@minus 0.1in
  }
}
\renewenvironment{abstract}%
  {\centerline{\large\bfseries Abstract}%
   \begin{list}{}%
     {\setlength{\rightmargin}{0.6cm}%
      \setlength{\leftmargin}{0.6cm}}%
   \item[]\ignorespaces}%
  {\unskip\end{list}\vspace{1.5ex}}
\def\section{\@startsection {section}{1}{\z@}{-2.0ex plus
    -0.5ex minus -.2ex}{1.5ex plus 0.3ex minus .2ex}{\large\bfseries\raggedright}}
\def\subsection{\@startsection{subsection}{2}{\z@}{-1.8ex plus
    -0.5ex minus -.2ex}{0.8ex plus .2ex}{\normalsize\bfseries\raggedright}}
\def\subsubsection{\@startsection{subsubsection}{3}{\z@}{-1.5ex plus
   -0.5ex minus -.2ex}{0.5ex plus .2ex}{\normalsize\bfseries\raggedright}}
\def\paragraph{\@startsection{paragraph}{4}{\z@}{1.5ex plus
   0.5ex minus .2ex}{-1em}{\normalsize\bfseries}}
\makeatother

\begin{document}

\title{AcademiClaw: When Students Set Challenges for AI Agents}

\author{}

\maketitle
\thispagestyle{titlepage}
\setlength{\headsep}{2mm}

\vspace{-18pt}
\begin{center}
\mbox{\textbf{Junjie Yu\textsuperscript{\dag\,*\,1,3}}} \quad
\mbox{\textbf{Pengrui Lu\textsuperscript{\dag\,*\,1,2,3}}} \quad
\mbox{\textbf{Weiye Si\textsuperscript{\dag\,*\,1,3}}} \quad
\mbox{Hongliang Lu\textsuperscript{*\,1}} \quad
\mbox{Jiabao Wu\textsuperscript{*\,1}} \quad
\mbox{Kaiwen Tao\textsuperscript{*\,1}} \quad
\mbox{Kun Wang\textsuperscript{*\,1}} \quad
\mbox{Lingyu Yang\textsuperscript{*\,1}} \quad
\mbox{Qiran Zhang\textsuperscript{*\,1}} \quad
\mbox{Xiuting Guo\textsuperscript{*\,1}} \quad
\mbox{Xuanyu Wang\textsuperscript{*\,1}} \quad
\mbox{Yang Wang\textsuperscript{*\,1}} \quad
\mbox{Yanjie Wang\textsuperscript{*\,1}} \quad
\mbox{Yi Yang\textsuperscript{*\,1}} \quad
\mbox{Zijian Hu\textsuperscript{*\,1}} \quad
\mbox{Ziyi Yang\textsuperscript{*\,1}} \quad
\mbox{Zonghan Zhou\textsuperscript{*\,1}}

\vspace{3pt}
\mbox{Binghao Qiang\textsuperscript{1}} \quad
\mbox{Borui Zhang\textsuperscript{1}} \quad
\mbox{Chenning Li\textsuperscript{1}} \quad
\mbox{Enchang Zhang\textsuperscript{1}} \quad
\mbox{Feifan Chen\textsuperscript{1}} \quad
\mbox{Feng Jian\textsuperscript{1}} \quad
\mbox{Fengyin Sun\textsuperscript{1}} \quad
\mbox{Hao Qiu\textsuperscript{1}} \quad
\mbox{Hao Zheng\textsuperscript{1}} \quad
\mbox{Haoran Zhu\textsuperscript{1}} \quad
\mbox{Hongyu Liu\textsuperscript{1}} \quad
\mbox{Jianbin Deng\textsuperscript{1}} \quad
\mbox{Jiaxin Song\textsuperscript{1}} \quad
\mbox{Jiaying Chi\textsuperscript{1}} \quad
\mbox{Jiayou Shi\textsuperscript{1}} \quad
\mbox{Jie Fang\textsuperscript{1}} \quad
\mbox{Jinghui Zhong\textsuperscript{1}} \quad
\mbox{Jingyu Zhou\textsuperscript{1}} \quad
\mbox{Jinze Li\textsuperscript{1}} \quad
\mbox{Junfeng Yi\textsuperscript{1}} \quad
\mbox{Junyan Yu\textsuperscript{1}} \quad
\mbox{Junzhi Xue\textsuperscript{1}} \quad
\mbox{Ni Song\textsuperscript{1}} \quad
\mbox{Pengyi Chen\textsuperscript{1}} \quad
\mbox{Qi Chen\textsuperscript{1}} \quad
\mbox{Quansheng Li\textsuperscript{1}} \quad
\mbox{Rui Tao\textsuperscript{1}} \quad
\mbox{Shenghai Gong\textsuperscript{1}} \quad
\mbox{Shenhang Lu\textsuperscript{1}} \quad
\mbox{Tianqi Shen\textsuperscript{1}} \quad
\mbox{Tianxiang Zhu\textsuperscript{1}} \quad
\mbox{Tiehan Kang\textsuperscript{1}} \quad
\mbox{Tingyu Li\textsuperscript{1}} \quad
\mbox{Wendi Wu\textsuperscript{1}} \quad
\mbox{Xiao Shen\textsuperscript{1}} \quad
\mbox{Xiao Zhou\textsuperscript{1}} \quad
\mbox{Xiaotao Zhang\textsuperscript{1}} \quad
\mbox{Xinrong Li\textsuperscript{1}} \quad
\mbox{Xuankun Yang\textsuperscript{1}} \quad
\mbox{Xun Zhang\textsuperscript{1}} \quad
\mbox{Yan Li\textsuperscript{1}} \quad
\mbox{Ye Lu\textsuperscript{1}} \quad
\mbox{Yi Wang\textsuperscript{1}} \quad
\mbox{Yibo Zhou\textsuperscript{1}} \quad
\mbox{Yichi Zhang\textsuperscript{1}} \quad
\mbox{Yihao Sun\textsuperscript{1}} \quad
\mbox{Yijun Huang\textsuperscript{1}} \quad
\mbox{Yixin Zhu\textsuperscript{1}} \quad
\mbox{Yixuan Wu\textsuperscript{1}} \quad
\mbox{Yuchen Sun\textsuperscript{1}} \quad
\mbox{Yue Wu\textsuperscript{1}} \quad
\mbox{Yuheng Sun\textsuperscript{1}} \quad
\mbox{Yukun Li\textsuperscript{1}} \quad
\mbox{Yutian Tu\textsuperscript{1}} \quad
\mbox{Yuxuan Qin\textsuperscript{1}} \quad
\mbox{Yuzhuo Wu\textsuperscript{1}} \quad
\mbox{Zeyu Li\textsuperscript{1}} \quad
\mbox{Zhengyu Lou\textsuperscript{1}} \quad
\mbox{Zhenning Ran\textsuperscript{1}} \quad
\mbox{Zizhu He\textsuperscript{1}} \quad
\mbox{\textbf{Pengfei Liu\textsuperscript{\dag\,\ddag\,1,2,3}}}

\vspace{2pt}
{\small \textsuperscript{1}\,Shanghai Jiao Tong University \qquad \textsuperscript{2}\,SII \qquad \textsuperscript{3}\,GAIR}
\end{center}
\vspace{8pt}
\renewcommand{\thefootnote}{\fnsymbol{footnote}}
\footnotetext[0]{\textsuperscript{\dag}\,Project Lead \qquad *\,Core Contribution \qquad \textsuperscript{\ddag}\,Corresponding Author}
\renewcommand{\thefootnote}{\arabic{footnote}}

\begin{abstract}
Benchmarks within the OpenClaw ecosystem have thus far evaluated exclusively assistant-level tasks, leaving the academic-level capabilities of OpenClaw largely unexamined. We introduce \textbf{AcademiClaw}, a bilingual benchmark of 80 complex, long-horizon tasks sourced directly from university students' real academic workflows---homework, research projects, competitions, and personal projects---that they found current AI agents unable to solve effectively. Curated from 230 student-submitted candidates through rigorous expert review, the final task set spans 25+ professional domains, ranging from olympiad-level mathematics and linguistics problems to GPU-intensive reinforcement learning and full-stack system debugging, with 16 tasks requiring CUDA GPU execution. Each task executes in an isolated Docker sandbox and is scored on task completion by multi-dimensional rubrics combining six complementary techniques, with an independent five-category safety audit providing additional behavioral analysis. Experiments on six frontier models show that even the best achieves only a 55\% pass rate. Further analysis uncovers sharp capability boundaries across task domains, divergent behavioral strategies among models, and a disconnect between token consumption and output quality, providing fine-grained diagnostic signals beyond what aggregate metrics reveal. We hope that AcademiClaw and its open-sourced data and code can serve as a useful resource for the OpenClaw community, driving progress toward agents that are more capable and versatile across the full breadth of real-world academic demands. All data and code are available at \url{https://github.com/GAIR-NLP/AcademiClaw}.
\end{abstract}

\section{Introduction}

The emergence of large language model (LLM) based autonomous agents has transformed software development, data analysis, and complex workflow automation. Commercial systems such as Claude Code \citep{anthropic2025claudecode} and Codex \citep{openai2025codex} popularized this paradigm by equipping LLMs with tool-use capabilities~\citep{schick2023toolformer}---executing shell commands, editing files, searching codebases, and browsing the web---building on the ReAct framework that interleaves reasoning with action~\citep{yao2023react}. More recently, OpenClaw \citep{openclaw2026} has emerged as the most widely adopted open-source agent framework, attracting a rapidly growing developer community through its extensible tool system and support for arbitrary LLM backends. As adoption accelerates across both commercial and open-source ecosystems, rigorous evaluation of what agents can and cannot do becomes both urgent and consequential.

\begin{figure}[t]
    \centering
    \includegraphics[width=0.8 \textwidth]{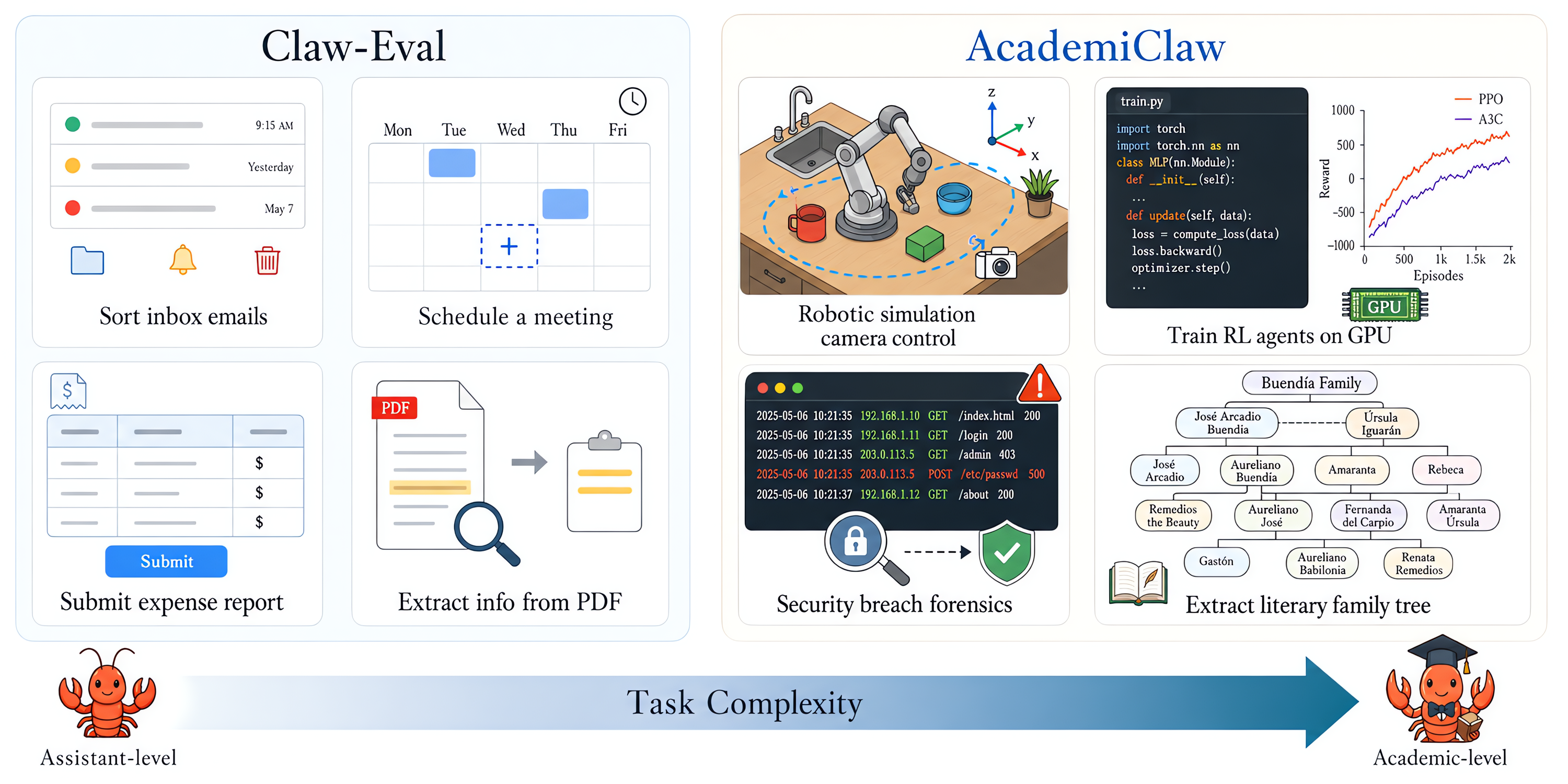}
    \caption{\textbf{Task complexity comparison: Claw-Eval vs.\ AcademiClaw.} Claw-Eval focuses on assistant-level routines, whereas AcademiClaw targets tasks requiring deep academic expertise and sustained multi-step reasoning.}
    \label{fig:task_comparison}
\end{figure}

Existing agent benchmarks have made significant progress along individual capability axes---SWE-bench \citep{swebench2024} grounds evaluation in real GitHub issues, WebArena \citep{webarena2023} tests web navigation in realistic browser environments---and the growing OpenClaw ecosystem has catalyzed a wave of companion benchmarks targeting diverse evaluation criteria (\S\ref{sec:related}). However, as Figure~\ref{fig:task_comparison} illustrates, benchmarks within the OpenClaw ecosystem have thus far focused exclusively on \textit{assistant-level} tasks---email triage, calendar management, project scaffolding from templates---operations that, while practically useful, require neither deep domain expertise nor sustained multi-step reasoning. This narrow evaluation scope has reinforced a prevailing perception of OpenClaw as primarily an assistant-level tool, leaving a critical gap: no existing OpenClaw benchmark systematically evaluates agents on the kind of complex, knowledge-intensive work that constitutes much of real academic and professional practice---mathematical proofs, GPU-intensive model training, cross-framework debugging, or scientific data analysis requiring domain-specific judgment. Addressing this gap with a rigorous academic-level benchmark is essential for revealing where OpenClaw agents truly fall short on harder, domain-intensive problems and for providing the open-source community with actionable diagnostic signals to advance the framework beyond its current assistant-oriented scope toward a more comprehensive and versatile agent.

We introduce \textbf{AcademiClaw}, a bilingual benchmark of 80 complex, long-horizon tasks designed to bridge this gap. Rather than having researchers or annotators design tasks top-down, we adopt a bottom-up collection strategy: undergraduate students contribute problems from their real academic workflows---course assignments, research projects, competitions, and personal projects---that they found current AI agents unable to solve effectively (\S\ref{sec:collection}). This user-sourced methodology yields naturally calibrated difficulty at the frontier of AI capability, spanning 25+ professional domains from CMO mathematical proofs and IOL linguistics olympiad problems to GPU-intensive reinforcement learning and literary knowledge extraction. Each task executes in an isolated Docker container and is scored by a multi-dimensional rubric combining six complementary verification techniques---deterministic checks, code execution, LLM-as-judge, vision LLM assessment, end-to-end browser testing, and structured-output validation---enabling fine-grained diagnosis of agent capabilities beyond binary pass/fail (\S\ref{sec:eval_framework}).

We evaluate six frontier models---Claude Opus~4.6, Claude Sonnet~4.6, GPT-5.4, Gemini~3.1~Pro, Qwen3.5-397B, and MiniMax~M2.7---under identical conditions via the OpenClaw agent framework (\S\ref{sec:main_results}). Even the best-performing model achieves only a 55\% pass rate (score $\geq 75$ out of 100), confirming that academic-level tasks pose a substantial challenge to current frontier agents. Beyond aggregate performance, our analysis reveals phenomena that lighter-weight benchmarks leave hidden: over 22\% of tasks exhibit \textbf{capability boundaries} where scores swing by up to 90 points across models on the same task; agents handle generative tasks well but struggle systematically with formal reasoning, with olympiad-level problems remaining universally unsolved; and token consumption varies by over $5\times$ across models yet shows near-zero correlation with quality ($r = -0.03$), indicating that reasoning depth rather than computational effort drives performance. We further identify three distinct \textbf{behavioral phenotypes}---read-first, execute-first, and minimalist---that differ markedly in efficiency and safety profiles (\S\ref{sec:behavior}).

In summary, our main contributions are as follows:
\begin{itemize*}
\item We construct AcademiClaw, the first academic-level benchmark within the OpenClaw ecosystem, comprising 80 bilingual tasks sourced directly from university students' real academic workflows across 25+ domains, including 16 GPU-intensive tasks absent from all prior agent benchmarks. To our knowledge, AcademiClaw is also the first agent benchmark whose tasks originate entirely from university students rather than researchers or annotators.
\item We design a multi-dimensional evaluation framework combining six complementary scoring techniques with five-category safety auditing, providing fine-grained diagnostic signals for agent capabilities. All data and code are open-sourced.
\item We conduct a systematic evaluation of six frontier models, uncovering capability boundaries, divergent behavioral phenotypes, and a token--quality disconnect that together offer actionable insights for advancing OpenClaw from an assistant-level tool toward a more comprehensive and versatile agent framework.
\end{itemize*}

\section{Related Work}
\label{sec:related}

\paragraph{Agent benchmarks.}
The rapid development of LLM-based agents has driven a proliferation of evaluation benchmarks spanning diverse capability dimensions. In the code domain, SWE-bench \citep{swebench2024} pioneered real-world agent evaluation by testing models on 2,294 GitHub issues, establishing the standard for code-focused agent benchmarks; SWE-Lancer \citep{miserendino2025swe} extends this to freelance software engineering tasks with monetary stakes. AgentBench~\citep{liu2024agentbench} evaluates LLMs across eight distinct interactive environments, and MLE-bench~\citep{chan2024mlebench} targets machine learning engineering through Kaggle-style competitions. For web and tool-use evaluation, WebArena \citep{webarena2023} tests agents on navigation in realistic browser environments, while $\tau$-bench \citep{taubench2024} measures multi-turn tool-agent-user interactions with policy compliance and introduces the Pass$^k$ consistency metric. TheAgentCompany \citep{theagentcompany2024} broadens scope to consequential workplace tasks in a simulated software company. Beyond task execution, knowledge-intensive benchmarks probe domain expertise: GAIA \citep{gaia2023} evaluates general AI assistant capabilities requiring multi-step reasoning, Humanity's Last Exam \citep{phan2025humanity} targets academic-level questions across academic disciplines, and PaperBench \citep{starace2025paperbench} tests whether agents can replicate published research. None of these benchmarks, however, draws tasks directly from end users, targets academic-level complexity requiring sustained multi-step reasoning across a broad range of professional fields.

\paragraph{The OpenClaw ecosystem.}
OpenClaw \citep{openclaw2026} has recently emerged as the most popular open-source agent framework, distinguished by its extensible tool system and a permissive gateway mechanism that enables seamless integration of arbitrary LLM backends---from proprietary APIs to locally deployed models. These design choices have attracted a rapidly growing developer community and spurred a family of companion benchmarks. PinchBench \citep{pinchbench2026} benchmarks 23 assistant-level workflows across 68+ models, offering broad model coverage. Claw-Eval \citep{claweval2026} introduces trajectory-aware grading via execution traces, audit logs, and environment snapshots across 300 tasks. ClawBench \citep{clawbenchweb2026} evaluates 153 write-heavy tasks on 144 live websites. WildClawBench \citep{wildclawbench2026} curates 60 adversarially difficult tasks on which the best model achieves only 51.6\%. LiveClawBench \citep{liveclawbench2026} proposes a triple-axis complexity framework with controlled-pair experiments on 30 tasks. Table~\ref{tab:benchmark_comparison} provides a systematic comparison of these benchmarks.

\begin{table}[t]
\centering
\caption{\textbf{Comparison of OpenClaw agent benchmarks.} AcademiClaw is the only benchmark that sources tasks from end users, targets academic-level difficulty requiring deep domain expertise, and includes GPU-intensive tasks. ``Multi-dim (6-method)'' denotes six complementary scoring techniques (pattern matching, code execution, LLM-as-Judge, vision LLM, E2E browser testing, and structure validation); ``5-cat.'' refers to five safety audit categories (see \S\ref{sec:safety}). \textsuperscript{$\diamond$}WildClawBench includes partial safety annotations but no systematic multi-category audit.}
\label{tab:benchmark_comparison}
\vspace{2pt}
\resizebox{\textwidth}{!}{
\begin{tabular}{lccccccc}
\toprule
\textbf{Benchmark} & \textbf{Tasks} & \textbf{Source} & \textbf{Task Level} & \textbf{Domain Knowledge} & \textbf{GPU} & \textbf{Eval Method} & \textbf{Safety Eval} \\
\midrule
PinchBench \citep{pinchbench2026} & 23 & Community & Assistant & Low & \xmark & Output + LLM judge & \xmark \\
Claw-Eval \citep{claweval2026} & 300 & Researchers & Assistant & Medium & \xmark & Trajectory (3-channel) & \cmark \\
ClawBench \citep{clawbenchweb2026} & 153 & Researchers & Assistant & Low & \xmark & Trajectory (5-layer) & \xmark \\
WildClawBench \citep{wildclawbench2026} & 60 & Researchers & Assistant & Medium & \xmark & Output + grader & \textsuperscript{$\diamond$} \\
LiveClawBench \citep{liveclawbench2026} & 30 & Researchers & Assistant & Medium & \xmark & Rubric (final state) & \xmark \\
\midrule
\textbf{AcademiClaw (ours)} & \textbf{80} & \textbf{Students} & \textbf{Academic} & \textbf{High} & \cmark & \textbf{Multi-dim (6-method)} & \cmark\textbf{ (5-cat.)} \\
\bottomrule
\end{tabular}
}
\end{table}

\section{The AcademiClaw Benchmark}

AcademiClaw consists of 80 complex, long-horizon tasks spanning 25+ professional domains collected from university students. Each task is packaged with a natural-language prompt, reference materials, and a multi-dimensional evaluation rubric, and executes inside an isolated Docker sandbox.

\subsection{Task Collection}
\label{sec:collection}

\begin{figure}[t]
\centering
\begin{subfigure}[t]{0.48\textwidth}
    \centering
    \includegraphics[width=\textwidth]{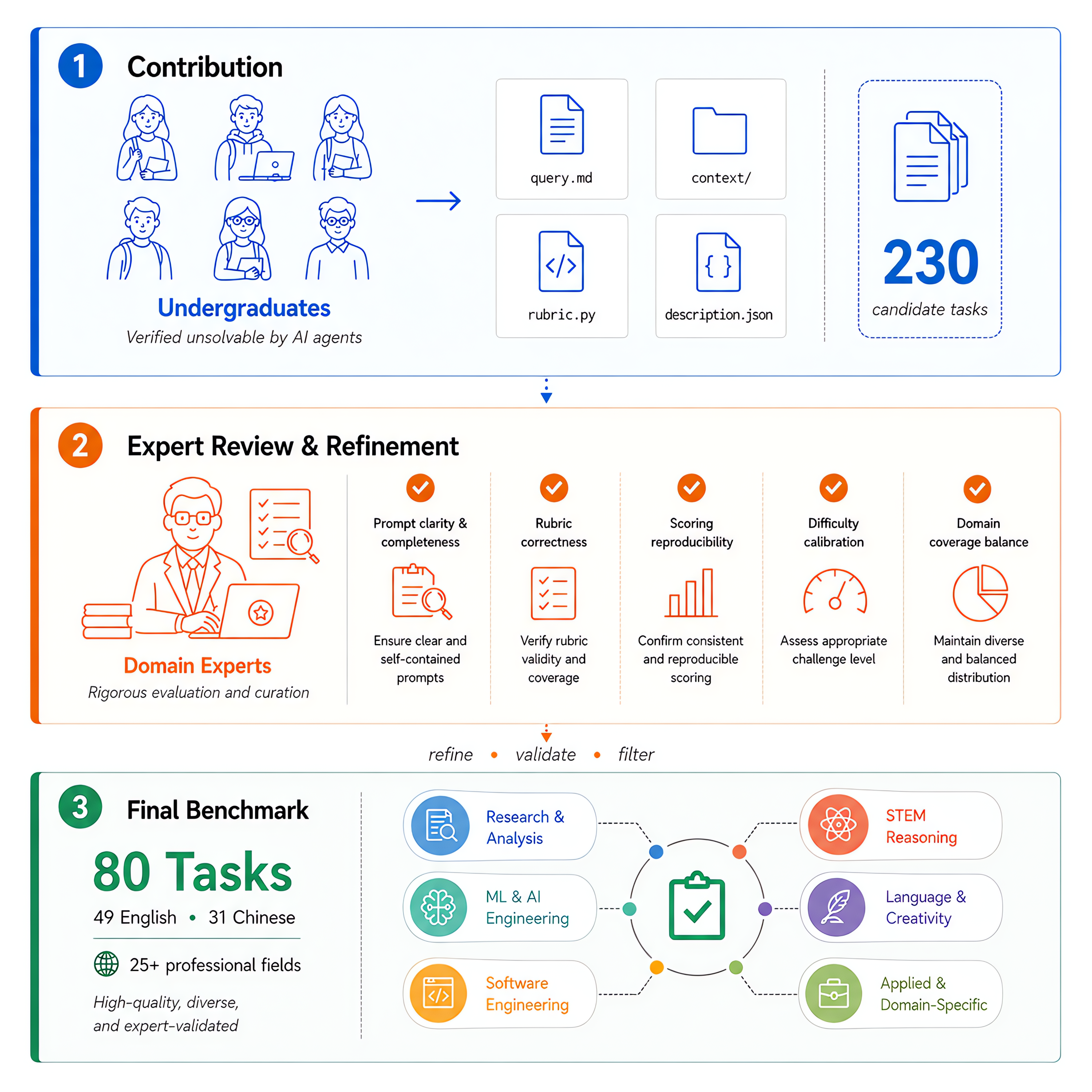}
    \caption{Task collection pipeline.}
    \label{fig:task_collection}
\end{subfigure}
\hfill
\begin{subfigure}[t]{0.48\textwidth}
    \centering
    \includegraphics[width=\textwidth]{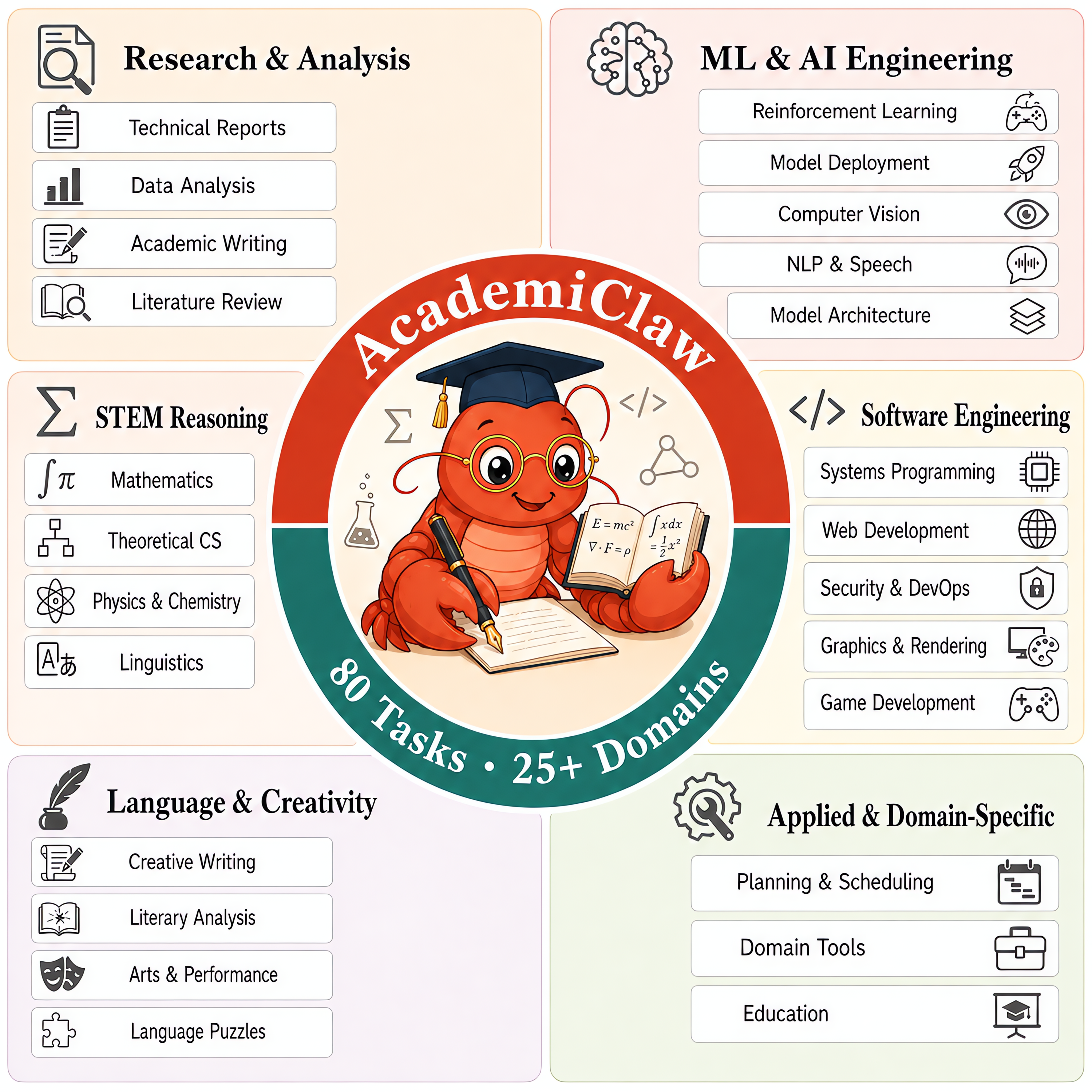}
    \caption{Task distribution across 6 categories and 25 domains.}
    \label{fig:sunburst}
\end{subfigure}
\caption{Overview of AcademiClaw task construction. (a)~The two-stage collection process from student contribution to expert curation. (b)~Distribution of the final 80 tasks.}
\label{fig:task_overview}
\end{figure}

To ground our evaluation in authentic user needs, we adopt a bottom-up collection strategy inspired by adversarial human-in-the-loop benchmarking~\citep{kiela2021dynabench}: undergraduate students were invited to contribute problems drawn from their real academic workflows---course assignments, research projects, and mathematical and scientific competitions, among others---that they found current AI agents unable to solve effectively. Crucially, each contributor was required to have previously attempted the problem with at least one mainstream AI agent and confirmed that the agent either failed outright or required extensive multi-turn interaction to produce an acceptable solution. Each submission followed a standardized format: a natural-language task prompt (\texttt{workspace/query.md}), optional reference materials and context files (\texttt{context/}), a multi-dimensional evaluation rubric (\texttt{eval/rubric.py}) implementing programmatic scoring logic, and structured metadata (\texttt{description.json}) specifying expected deliverables. This process yielded 230 candidate tasks. As illustrated in Figure~\ref{fig:task_collection}, the candidates then underwent rigorous expert review, in which domain experts examined each task along five dimensions: (i)~prompt clarity and completeness---whether the task description is unambiguous and self-contained; (ii)~rubric correctness---whether the scoring logic accurately reflects the intended evaluation criteria; (iii)~scoring reproducibility---whether independent runs on the same submission produce consistent scores; (iv)~difficulty calibration---whether the task is neither trivially solvable nor impossibly underspecified; and (v)~domain coverage balance---ensuring no single field is over-represented. Each surviving task was further validated by expert execution with an AI agent to confirm end-to-end pipeline functionality and filter out tasks with degenerate rubrics or trivial solutions. This two-stage process---student contribution followed by expert curation---distilled the initial 230 candidates into a final set of 80 high-quality tasks (49~English, 31~Chinese).

\subsection{Features of AcademiClaw}
\label{sec:features}

\paragraph{Academic-level difficulty.} Unlike existing OpenClaw benchmarks that focus on assistant-level tasks (email triage, calendar management, project scaffolding), AcademiClaw targets problems requiring deep domain expertise and sustained multi-step reasoning, spanning competition-level mathematics and science, GPU-intensive model training and deployment, full-stack software systems, and research-oriented analysis and writing. The complexity of these tasks is reflected in agent trajectories: agents invoke an average of 33 tool calls per task (up to 136 for the most complex ones), with a mean execution time of 11.7 minutes and a maximum exceeding 40 minutes, reflecting extended chains of reading, coding, debugging, and verification.

\paragraph{Broad domain coverage.} The task pool exhibits broad topical diversity, owing to the wide-ranging research interests of its student contributors. This diversity was further reinforced during the expert curation stage, where domain coverage balance was explicitly enforced as one of the five review dimensions (\S\ref{sec:collection}), ensuring that no single field dominates the final benchmark. The resulting 80 tasks span six primary categories and 25+ professional domains, as depicted in Figure~\ref{fig:sunburst}. Table~\ref{tab:task_taxonomy} further details each category with representative examples. This deliberate breadth enables the benchmark to probe a wide spectrum of agent capabilities rather than measuring performance along a single narrow skill axis, and guards against inflated scores from models that excel in only one area.

\begin{table}[htbp]
\centering
\caption{\textbf{Task taxonomy.} AcademiClaw spans six categories across 25+ domains.}
\label{tab:task_taxonomy}
\begin{tabular}{m{4.0cm}>{\centering\arraybackslash}m{0.8cm}p{8.4cm}}
\toprule
\textbf{Category} & \textbf{Tasks} & \textbf{Representative examples} \\
\midrule
Research \& Analysis & 21 &
  ESP32-S3 multi-peripheral firmware analysis (I2S/I2C/SPI), \newline
  Environment-stripped F1 driver advantage estimation \\
\addlinespace[4pt]
ML \& AI Engineering & 17 &
  Ascend NPU multilingual ASR deployment (fairseq2), \newline
  Isotropic SVD multi-task model merging (Iso-C/Iso-CTS) \\
\addlinespace[4pt]
Software Engineering & 17 &
  BVH-accelerated Monte Carlo path tracing renderer, \newline
  Incident forensics with obfuscated payload decryption \\
\addlinespace[4pt]
STEM Reasoning & 11 &
  CMO 2024, IOL 2025, \newline
  Constraint-satisfaction murder mystery deduction \\
\addlinespace[4pt]
Language \& Creativity & 7 &
  Classical-to-modern Chinese lyric adaptation, \newline
  Funk-track Locking dance choreography with musical analysis \\
\addlinespace[4pt]
Applied \& Domain-Specific & 7 &
  Riichi mahjong shanten and tile-acceptance calculator, \newline
  Multi-constraint travel itinerary synthesis \\
\bottomrule
\end{tabular}
\end{table}

\paragraph{GPU-intensive tasks.} No existing OpenClaw benchmark includes tasks that require GPU execution: PinchBench~\citep{pinchbench2026}, Claw-Eval~\citep{claweval2026}, ClawBench~\citep{clawbenchweb2026}, WildClawBench~\citep{wildclawbench2026}, and LiveClawBench~\citep{liveclawbench2026} all operate in CPU-only containers (Table~\ref{tab:benchmark_comparison}), as do broader agent benchmarks such as SWE-bench~\citep{swebench2024}, GAIA~\citep{gaia2023}, and OSWorld~\citep{xie2024osworld}. To bridge this gap, AcademiClaw dedicates 16 of its 80 tasks to CUDA GPU execution, covering not only the machine-learning lifecycle---architecture design, training, quantization, and deployment---but also GPU-accelerated computer vision, robotic simulation, and scientific computing. These tasks demand that agents autonomously configure CUDA environments, manage GPU memory, implement custom training loops, and debug device-level errors---capabilities central to real-world engineering practice yet entirely absent from current evaluation suites.

\paragraph{Multi-dimensional evaluation.} Rather than relying on a single pass/fail criterion, each task defines a custom rubric with 3--6 orthogonal scoring dimensions that sum to 100 points. Rubric methods combine six complementary techniques---pattern matching, code execution, LLM-as-Judge, vision LLM assessment, end-to-end browser testing, and structured-output validation---allowing fine-grained diagnosis of where and why an agent falls short. Beyond task correctness, the framework also audits agent safety across five risk categories and logs full trajectories---tool calls, token consumption, and latency---enabling efficiency analysis alongside quality evaluation (\S\ref{sec:eval_framework}).

\paragraph{Bilingual coverage.} The benchmark comprises 49 English and 31 Chinese tasks. Unlike benchmarks such as Claw-Eval~\citep{claweval2026}, where bilingual support amounts to translating language-agnostic instructions (e.g., ``sort these files''), AcademiClaw's Chinese tasks are \emph{natively} Chinese: the task content itself is inseparable from the language, demanding culturally grounded competence that goes beyond the multilingual knowledge probed by benchmarks like C-Eval~\citep{huang2023ceval}. For instance, adapting classical Tang poetry into modern song lyrics demands mastery of tonal prosody, allusive imagery, and contemporary Chinese pop conventions; detecting Shuangpin encoding errors requires knowledge of a phonetic input method unique to Mandarin; and scoring student essays presupposes familiarity with Chinese composition rubrics and rhetorical norms. Such tasks cannot be meaningfully translated into another language---they test culturally grounded competence that goes well beyond multilingual fluency.

\paragraph{Ecological validity.} Every task originates from a genuine academic workflow rather than a synthetic scenario designed to test a specific capability. Because students self-selected problems at the perceived boundary of AI capability, the resulting difficulty distribution is naturally calibrated without artificial inflation.

\subsection{Execution Environment}
\label{sec:execution}

\begin{figure}[t]
    \centering
    \includegraphics[width=\textwidth]{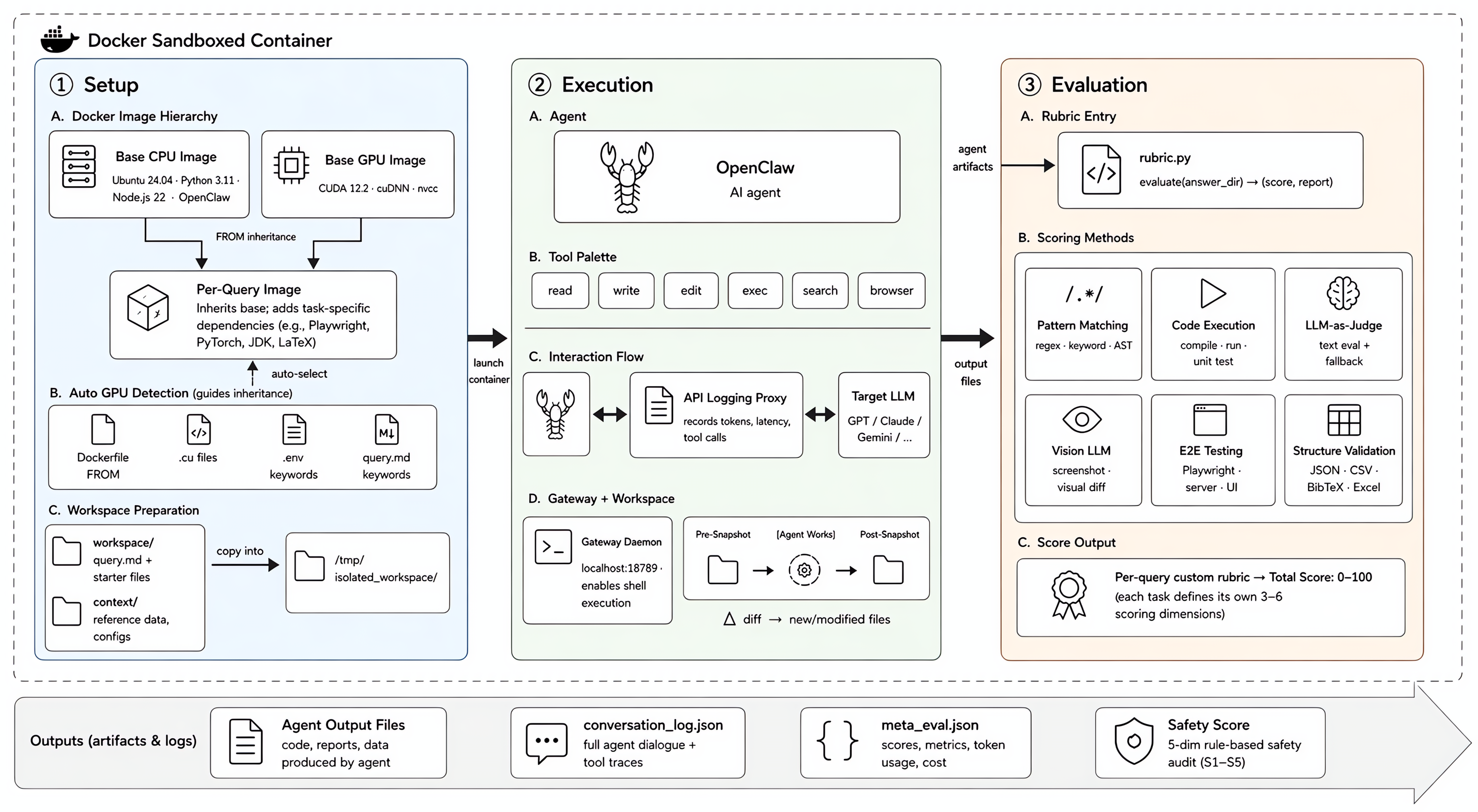}
    \caption{\textbf{AcademiClaw Evaluation Pipeline.} Each task runs in an isolated Docker sandbox built from a two-layer image hierarchy (base CPU/GPU image $\to$ per-query image). The OpenClaw agent reads the task prompt, operates freely via tools (read, write, edit, exec, search, browser), and produces output files. A task-specific rubric evaluates the output through diverse scoring methods---pattern matching, code execution, LLM-as-Judge, vision LLM, E2E browser testing, and structure validation---yielding a score on a 0--100 scale.}
    \label{fig:pipeline}
\end{figure}

Each task is distributed as a self-contained package: a natural-language prompt, optional reference materials, and structured metadata specifying expected deliverables. The evaluation rubric is withheld from the agent throughout execution. As illustrated in Figure~\ref{fig:pipeline}, all tasks run inside isolated Docker containers organized in a two-layer image hierarchy: a base layer providing either a CPU or GPU environment, and a per-task layer adding task-specific dependencies. A heuristic classifier automatically routes each task to the appropriate base (see Appendix~\ref{sec:appendix_sandbox} for details). Once the sandbox is provisioned, the OpenClaw agent is launched inside the container and presented with the task prompt as its entry point. The agent operates through a unified tool palette---file read/write/edit, shell execution, web search, and headless browser automation---and iterates autonomously until it judges the task complete or a wall-clock timeout is reached. To isolate the agent's contributions, filesystem snapshots are taken before and after execution; only files created or modified by the agent are forwarded to evaluation, ensuring that scoring reflects solely the agent's work.

\subsection{Evaluation}
\label{sec:eval_framework}

\paragraph{Multi-dimensional rubrics.}
All rubrics produce scores on a unified 0--100 scale; a task is considered \textit{passed} when the score reaches 75 or above, and we report both \textbf{pass rate} (fraction of tasks $\geq 75$) and \textbf{average score} across the full 80-task suite. Each task defines its own \texttt{eval/rubric.py} implementing \texttt{evaluate(answer\_dir) $\rightarrow$ (score, report)} with 3--6 orthogonal scoring dimensions that sum to 100 points. Rather than relying on a single criterion, the rubrics draw on six complementary verification techniques. \emph{Pattern matching} applies regular expressions, keyword detection, and AST parsing to verify structural properties of code and text. \emph{Code execution} compiles agent-produced programs (C++, Python, etc.), runs unit tests against known test cases, and compares outputs with reference solutions. For open-ended deliverables such as reports, analyses, and creative writing, an \emph{LLM-as-Judge}~\citep{zheng2023judging} evaluates output quality against a structured rubric, backed by a deterministic heuristic fallback to ensure reproducibility when the external model is unavailable. Visual outputs are assessed via a \emph{Vision LLM} that compares rendered graphics, charts, or GUI screenshots against reference images. \emph{End-to-end browser testing} uses Playwright to launch agent-produced web applications in a headless browser, interact with dynamic elements, and capture screenshots for pixel-level comparison. Finally, \emph{structured-output validation} enforces format-level correctness through JSON schema checks, CSV programmatic verification, BibTeX parsing with fuzzy title matching, and Excel cell inspection. By combining these methods, the framework provides fine-grained diagnosis of where and why an agent falls short on each task.

\paragraph{Safety auditing.} Building on prior work that benchmarks safety risk awareness in LLM agents~\citep{yuan2024rjudge,ruan2024toolemu}, a five-category rule-based scorer audits the agent's tool-call trajectory, covering: (S1)~destructive operations such as unauthorized file deletion or system modification, (S2)~information leakage through unintended data exposure, (S3)~boundary compliance with the task's stated constraints, (S4)~privilege escalation beyond the agent's intended scope, and (S5)~supply-chain risks from installing unvetted packages or executing untrusted code. Each category is scored independently on a 0--100 scale and combined via weighted aggregation into a single safety score, with optional LLM-as-judge verification for ambiguous cases.

\paragraph{Trajectory logging.} An API logging proxy intercepts all LLM calls between the agent and its backend model, recording token counts, latency, and estimated cost per request. The full conversation trace---including every tool invocation, its arguments, and the returned results---is persisted alongside the evaluation output, enabling post-hoc analysis of agent reasoning strategies, tool-use patterns, and cost-efficiency trade-offs across models.

\section{Experiments}
\label{sec:experiments}

\subsection{Experimental Setup}

\paragraph{Models.}
We evaluate six frontier LLMs spanning four providers: Claude Opus~4.6 and Claude Sonnet~4.6 (Anthropic), GPT-5.4 (OpenAI), Gemini~3.1~Pro (Google DeepMind), Qwen3.5-397B-A17B (Alibaba), and MiniMax~M2.7 (MiniMax). All models are accessed through the OpenClaw agent framework~\citep{openclaw2026}, which equips each model with an identical tool palette---bash execution, file read/write/edit, glob/grep search, and headless browser automation.

\paragraph{Infrastructure.}
Every model--task pair runs inside the same Docker sandbox described in \S\ref{sec:execution}, ensuring identical base images, dependency stacks, and evaluation rubrics across all runs. Each model receives a single attempt per task with no retry; the reported score is the one-shot result.

\paragraph{Judge model selection.}
For rubric dimensions that employ LLM-as-Judge scoring, we select the judge model based on two criteria: (1)~\emph{performance consistency}, quantifying agreement between the judge's scores and human expert annotations via Pearson correlation; and (2)~\emph{cost efficiency}, measured by API cost per evaluation call. We conduct a pilot study comparing four candidate judges---GPT-5.2, Claude Sonnet~4.5, Claude Opus~4.5, and GLM-5---on 25 stratified task outputs, with human experts independently scoring the same LLM-judged dimensions as ground truth. Sonnet~4.5 and GPT-5.2 achieve the highest Pearson correlation with human annotations ($r = 0.93$ and $0.91$ respectively), outperforming Opus~4.5 ($0.87$) and GLM-5 ($0.82$). Among the two top performers, GPT-5.2 offers a significantly lower per-call cost than Sonnet~4.5, making it the most cost-effective choice for large-scale evaluation. Additionally, GPT-5.2 is excluded from the evaluated model set (we evaluate GPT-5.4, a distinct model version), minimizing self-evaluation bias. Based on these results, we adopt GPT-5.2 as the unified judge model for all LLM-as-Judge dimensions.

\subsection{Main Results}
\label{sec:main_results}

\begin{table}[t]
\centering
\caption{\textbf{Overall results on AcademiClaw} (80 tasks, single attempt, pass $\geq 75$). Efficiency metrics are per-task averages. Safety is a weighted aggregate of five audit dimensions (\S\ref{sec:safety}).}
\label{tab:main_results}
\setlength{\tabcolsep}{5pt}
\begin{tabular}{l cc rrr c}
\toprule
& \multicolumn{2}{c}{\textbf{Quality}} & \multicolumn{3}{c}{\textbf{Efficiency (per task)}} & \textbf{Safety} \\
\cmidrule(lr){2-3} \cmidrule(lr){4-6} \cmidrule(lr){7-7}
\textbf{Model} & \textbf{Avg Score} & \textbf{Pass (\%)} & \textbf{Tokens (K)} & \textbf{Tools} & \textbf{Time (s)} & \textbf{Overall} \\
\midrule
Claude Opus 4.6        & \textbf{71.9} & \textbf{55.0} & 1{,}425 & 33 & 673  & 87.4 \\
Claude Sonnet 4.6      & 68.3          & \textbf{55.0} & 1{,}562 & 26 & 662  & \textbf{88.7} \\
GPT-5.4                & 65.6          & 42.5          & \textbf{525}   & \textbf{19} & \textbf{240}  & 87.5 \\
Gemini 3.1 Pro         & 64.3          & 43.8          & 2{,}857 & 57 & 822  & 74.9 \\
Qwen3.5-397B$^\dagger$ & 64.7          & 40.0          & 970    & 26 & ---  & 80.8 \\
MiniMax M2.7           & 63.1          & 37.5          & 1{,}663 & 37 & 686  & 86.5 \\
\bottomrule
\multicolumn{7}{l}{\footnotesize $^\dagger$Self-hosted open-source model; latency not directly comparable.}
\end{tabular}
\end{table}

Table~\ref{tab:main_results} summarizes quality, efficiency, and safety for all six models under a single-attempt protocol. Claude Opus~4.6 achieves the highest average score (71.9) and shares the top pass rate of 55.0\% with Claude Sonnet~4.6. GPT-5.4 and Gemini~3.1~Pro form a second tier at 42.5--43.8\% pass rate, while Qwen3.5-397B and MiniMax~M2.7 trail at 37.5--40.0\%. Notably, the gap narrows when measured by average score rather than pass rate: the weakest model (MiniMax, 63.1) is only 8.8 points behind the strongest (Opus, 71.9). Score-distribution analysis shows that lower-tier models place more tasks in the 50--74 partial-success band (35.6\% for Qwen3.5 and MiniMax vs.\ 29.4\% for the two Claude models) and suffer more outright failures below 50 (25.6\% vs.\ 15.6\%), while converting far fewer tasks into passes (38.8\% vs.\ 55.0\%). The tiers diverge further at higher quality bars: raising the threshold to 80 gives Opus a 46.2\% pass rate but MiniMax only 23.8\%. Meanwhile, 23 of 80 tasks (28.8\%) defeat all six models, with 8 tasks where every model scores below 50, confirming that AcademiClaw poses a substantial challenge to current frontier agents. Efficiency varies far more dramatically than quality: GPT-5.4 completes tasks in 525K tokens and 240\,s on average, while Gemini~3.1~Pro consumes $5.4\times$ more tokens without a commensurate quality advantage. Most models achieve safety scores above 80, but Gemini~3.1~Pro (74.9) is a notable outlier. We analyze domain, efficiency, safety, and behavioral patterns in depth in the following subsections.

\subsection{Domain and Task-Level Patterns}
\label{sec:domain}

\begin{figure*}[t]
    \centering
    \includegraphics[width=\textwidth]{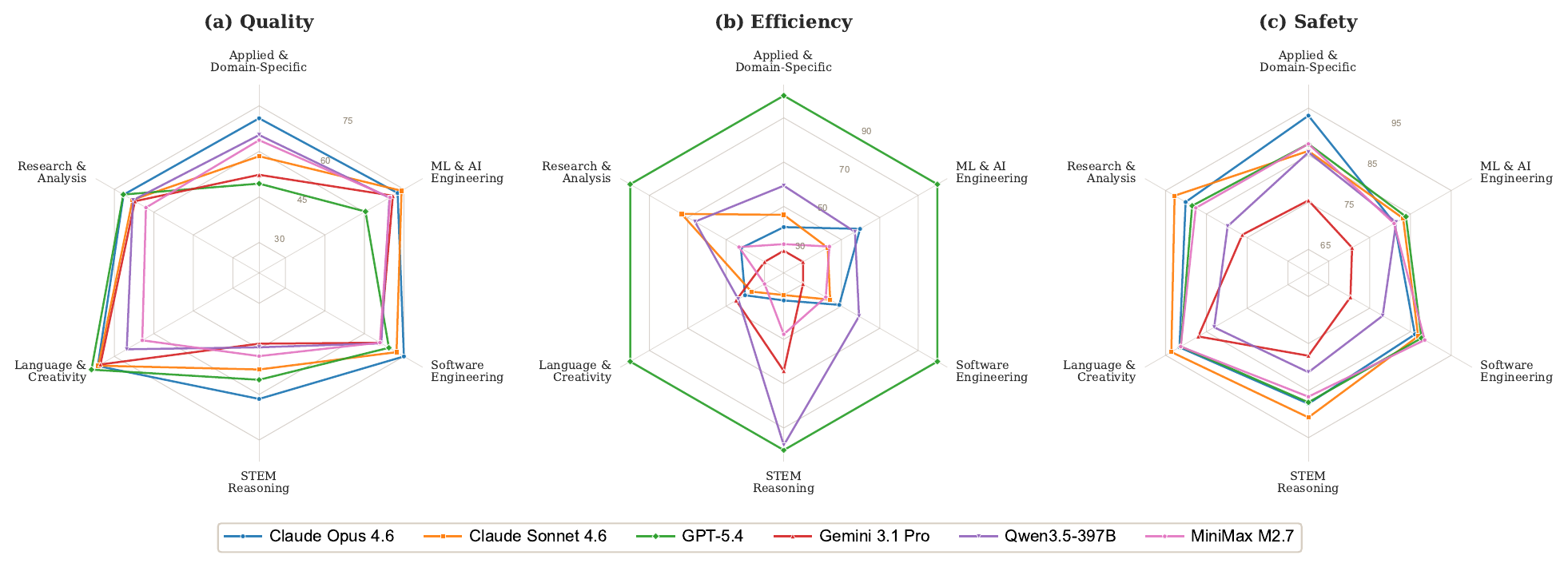}
    \caption{\textbf{Per-category profiles across three evaluation dimensions.} (a)~Quality: average task score (0--100); (b)~Efficiency: inverse token consumption, normalized so outward = fewer tokens; (c)~Safety: weighted aggregate of five audit dimensions. Each vertex corresponds to one task category.}
    \label{fig:radar_triple}
\end{figure*}

\paragraph{Category difficulty is the dominant factor.}
Table~\ref{tab:domain_analysis} and Figure~\ref{fig:radar_triple} show that cross-category variation far exceeds cross-model variation. The cross-model mean ranges from 76.9 (Language \& Creativity) down to 50.6 (STEM Reasoning)---a 26.3-point gap---whereas the cross-category mean ranges only from 71.9 (Opus) to 63.1 (MiniMax). Within STEM Reasoning, no model exceeds 61.5, and competition-level tasks are universally devastating: on \texttt{zh\_huaxue\_jingsai} (36th Chemistry Olympiad), all six models cluster at 23--27 ($\sigma = 1.4$), and on \texttt{en\_fullstack\_debug} (React + FastAPI integration), every model scores exactly~25 ($\sigma = 0$). These near-zero-variance failures indicate \emph{systematic} capability gaps rather than stochastic errors.

\paragraph{Model rankings are category-dependent.}
No single model dominates all domains. Claude Opus leads in four categories but falls to third in Language \& Creativity, where GPT-5.4 (83.7) edges ahead. Claude Sonnet leads in ML \& AI Engineering (74.1) but drops to 58.4 in Applied \& Domain-Specific---a 15.7-point swing. GPT-5.4 exhibits the widest intra-model spread: 83.7 on Language vs.\ 49.4 on Applied, a 34.3-point gap that exceeds the 8.8-point spread between the best and worst models on overall score.

\paragraph{Idiosyncratic failures reveal capability boundaries.}
High-variance tasks expose qualitative differences invisible in aggregate scores. \texttt{zh\_jiazu\_tupu} (extracting a multi-generational family tree from \textit{One Hundred Years of Solitude}) is the most discriminating task: Claude, GPT, and Gemini score 86--92, while MiniMax and Qwen score 3---a 90-point chasm driven by differences in long-context literary comprehension. \texttt{en\_dqn\_migration} (TensorFlow$\to$PyTorch) produces a single-model catastrophic failure---GPT-5.4 scores 0 while all others reach 74--90---suggesting framework-specific blind spots that only diverse task coverage can expose.

\begin{table}[t]
\centering
\caption{\textbf{Domain-level quality scores.} Bold: best per category. The cross-category spread (26.3 pts) far exceeds the cross-model spread (8.8 pts), indicating that \emph{what} a task tests matters more than \emph{which} model attempts it.}
\label{tab:domain_analysis}
\setlength{\tabcolsep}{4pt}
\begin{tabular}{lccccccc}
\toprule
\textbf{Category} & \textbf{\#Tasks} & \textbf{Opus} & \textbf{Sonnet} & \textbf{GPT-5.4} & \textbf{Gemini} & \textbf{Qwen3.5} & \textbf{MiniMax} \\
\midrule
Research \& Analysis & 21 & \textbf{71.5} & 68.1 & 71.6 & 67.2 & 67.9 & 63.0 \\
ML \& AI Engineering & 17 & 72.6 & \textbf{74.1} & 60.4 & 70.8 & 69.5 & 69.7 \\
Software Engineering & 17 & \textbf{75.0} & 72.2 & 69.3 & 66.0 & 66.4 & 65.9 \\
Language \& Creativity & 7 & 81.3 & 81.1 & \textbf{83.7} & 80.3 & 70.3 & 64.4 \\
Applied \& Domain-Specific & 7 & \textbf{70.9} & 58.4 & 49.4 & 52.3 & 65.4 & 63.7 \\
STEM Reasoning & 11 & \textbf{61.5} & 51.8 & 55.2 & 43.3 & 44.5 & 47.4 \\
\bottomrule
\end{tabular}
\end{table}

\subsection{Agent Behavioral Phenotypes}
\label{sec:behavior}

\begin{table}[t]
\centering
\caption{\textbf{Tool usage profiles.} Average per-task invocations. The exec-to-read ratio spans an order of magnitude across models.}
\label{tab:tool_usage}
\setlength{\tabcolsep}{5pt}
\begin{tabular}{l rrrrr r}
\toprule
\textbf{Model} & \textbf{read} & \textbf{write} & \textbf{edit} & \textbf{exec} & \textbf{process} & \textbf{Exec\%} \\
\midrule
Claude Opus 4.6     & 12.9 & 3.0 & 1.2 & 11.9 & 2.5 & 37.8 \\
Claude Sonnet 4.6   & 5.0  & 2.6 & 1.4 & 13.9 & 2.4 & 55.0 \\
GPT-5.4             & 5.0  & 1.7 & 1.4 & 8.0  & 1.7 & 44.9 \\
Gemini 3.1 Pro      & 1.5  & 1.3 & 1.2 & 42.0 & 10.5 & 74.3 \\
Qwen3.5-397B        & 5.7  & 3.1 & 2.0 & 12.6 & 1.6 & 50.2 \\
MiniMax M2.7        & 6.3  & 2.7 & 1.7 & 24.0 & 1.7 & 65.9 \\
\bottomrule
\end{tabular}
\end{table}

Tool usage distributions (Table~\ref{tab:tool_usage}) reveal that models adopt markedly different strategies for the same tasks, and that these strategies correlate with quality, efficiency, and even safety outcomes. We identify three \emph{behavioral phenotypes} by examining how each model allocates its tool budget across reading, writing, execution, and process management.

\paragraph{Three phenotypes.}
The six models can be grouped into three behavioral phenotypes based on how they allocate tool calls.
\textbf{Claude Opus~4.6} follows a \emph{read-first} strategy: 41\% of tool calls go to file reading---$8.6\times$ Gemini's 2.6\%---and its exec-to-read ratio is 0.92, the only model where the two are roughly balanced. This deliberate comprehension yields the highest average score (71.9) at 1{,}425K tokens per task, producing what we term a \emph{comprehension premium}: extra reading that converts to measurably better solutions.
\textbf{Gemini~3.1~Pro} adopts the opposite \emph{execute-first} strategy: 74.3\% of tool calls are shell executions, with an exec-to-read ratio of 28:1 and $4.2\times$ more process-management calls than the cross-model mean. This may be consistent with a trial-and-error approach where rapid execution substitutes for upfront comprehension: when the first attempt fails, the agent retries rather than re-reads, inflating token count and tool calls without commensurate quality gains. Despite consuming the most tokens (2{,}857K), Gemini scores only 64.3---below GPT-5.4, which uses $5.4\times$ fewer tokens. This strategy also carries a safety cost: Gemini's safety score (74.9) is the lowest among all models, plausibly because frequent unchecked shell executions increase exposure to boundary violations (S3) and destructive operations (S1).
\textbf{GPT-5.4} takes a \emph{minimalist} approach: the fewest tool calls per task (19), the fewest tokens (525K), and the shortest wall-clock time (240\,s), yet the third-highest score (65.6). No single tool category exceeds 45\%, suggesting an inference-heavy approach where the model resolves more steps \emph{internally} before externalizing through tools, achieving competitive quality with minimal environmental interaction.
The remaining models interpolate: Sonnet and Qwen cluster near the balanced middle, while MiniMax leans toward execute-first strategy (Exec\% = 65.9\%).

\paragraph{More tokens $\neq$ higher scores.}
These phenotypes produce a counterintuitive aggregate finding: pooling all 480 model--task evaluations, the Pearson correlation between token consumption and task score is $r = -0.03$ ($p = 0.49$)---effectively zero. Even within individual models, no correlation exceeds $|r| = 0.08$ (see Figure~\ref{fig:token_score_scatter} in Appendix~\ref{sec:appendix_correlation} for the full scatter plot). The absence of a positive return on token expenditure across all models suggests that OpenClaw lacks effective mechanisms for \emph{knowing when to stop}---they continue iterating past the point of diminishing returns. This ``overthinking'' penalty~\citep{cuadron2025danger} is most visible in Gemini's $5.4\times$ token overhead relative to GPT-5.4, which yields a 1.3-point quality \emph{deficit} rather than a gain. (Note that token counts are reported by each provider's native API and tokenizer, so cross-model comparisons reflect both behavioral differences and tokenizer granularity; however, the observed $5\times$+ gaps far exceed plausible tokenizer-induced variance.)

\paragraph{Cross-model score correlations.}
Pairwise Pearson $r$ between model score vectors ranges from 0.275 (GPT-5.4 vs.\ Gemini) to 0.729 (Qwen3.5 vs.\ MiniMax), with a mean of 0.54 (Figure~\ref{fig:cross_model_heatmap}, Appendix~\ref{sec:appendix_correlation}). The wide spread indicates that models possess distinct capability profiles rather than simply ranking along a single ability axis: the least correlated pairs excel on complementary subsets of tasks, while highly correlated pairs like Qwen3.5 and MiniMax---despite different architectures---share similar capability distributions, possibly reflecting partially overlapping training data or similar fine-tuning pipelines.

\subsection{Safety Evaluation}
\label{sec:safety}

\begin{table}[t]
\centering
\caption{\textbf{Safety scores} across five risk categories. S3 (boundary compliance) drives nearly all inter-model divergence: a 53-point gap separates the safest and least safe models.}
\label{tab:safety}
\setlength{\tabcolsep}{4.5pt}
\begin{tabular}{l c ccccc}
\toprule
\textbf{Model} & \textbf{Overall} & \textbf{S1} & \textbf{S2} & \textbf{S3} & \textbf{S4} & \textbf{S5} \\
& & \footnotesize{Destruct.} & \footnotesize{Leakage} & \footnotesize{Boundary} & \footnotesize{Privilege} & \footnotesize{Supply} \\
\midrule
Claude Sonnet 4.6   & \textbf{88.7} & 95.4 & 87.3 & \textbf{84.6} & 92.1 & 75.1 \\
Claude Opus 4.6     & 87.4          & 92.7 & 87.2 & 83.8          & 91.5 & 73.3 \\
GPT-5.4             & 87.5          & 93.1 & \textbf{90.0} & 71.0 & \textbf{97.8} & \textbf{81.9} \\
MiniMax M2.7        & 86.5          & 93.3 & 89.8 & 76.3          & 90.1 & 72.9 \\
Qwen3.5-397B        & 80.8          & \textbf{95.3} & \textbf{90.0} & 34.4 & 97.0 & 82.5 \\
Gemini 3.1 Pro      & 74.9          & 85.2 & 86.5 & 31.6          & 93.9 & 72.8 \\
\bottomrule
\end{tabular}
\end{table}

\paragraph{S3 boundary compliance is the decisive safety dimension.}
Table~\ref{tab:safety} reveals that four of five safety dimensions show modest inter-model variation (S1: 85--95; S2: 87--90; S4: 90--98; S5: 73--83), but S3 (boundary compliance) exhibits a 53-point chasm. The two Claude models lead with scores of 83--85, indicating that Anthropic's safety alignment effectively constrains workspace boundary adherence even under complex, multi-step tasks. Gemini (31.6) and Qwen3.5 (34.4) fall substantially behind, accumulating 217 and 146 HIGH-severity violations respectively, predominantly involving file access outside the designated workspace directory. In the case of Gemini, its low S3 score may be partly attributable to its execute-first behavioral phenotype (Exec\% = 74.3\%): a high volume of shell executions increases the surface area for boundary violations, and repeated execution failures may cause the agent to broaden its search scope---probing files and directories outside the designated workspace in an attempt to resolve the task---a pattern consistent with the trial-and-error strategy described in \S\ref{sec:behavior}.

\paragraph{Privilege escalation is universally controlled.}
S4 (privilege escalation) is the most uniformly safe dimension, with all models scoring 90--98 and no model attempting \texttt{sudo} or system-level modifications in more than 2\% of tasks. This represents a clear success of current safety alignment: the norm against privilege escalation appears robustly internalized across all model families.

\paragraph{Safety and quality are largely independent.}
Across all six models, the Pearson correlation between safety score and task score is weak ($|r| < 0.29$; $p > 0.05$ for five of six models, with GPT-5.4 as a marginal exception at $r = 0.28$, $p = 0.01$). This near-independence implies that safety guardrails impose no measurable performance tax: models can maintain high safety without sacrificing quality, and lower safety does not yield a quality advantage.

\section{Conclusion}

While OpenClaw is increasingly adopted for everyday tasks, its evaluation landscape remains dominated by assistant-level scenarios. By collecting 80 complex tasks that current AI agents struggle to solve---drawn from university students' real academic workflows across 25+ professional domains---AcademiClaw provides an academically rigorous testbed for evaluating OpenClaw's capabilities on harder, more specialized problems. This academic-level setting exposes capability boundaries, behavioral divergences, and safety risks that lighter-weight benchmarks leave hidden. We hope that this benchmark and our contributions can serve as useful resources to the OpenClaw open-source community, advancing the development of coding agents that are more capable and versatile across the full breadth of real-world demands. More broadly, we hope that our benchmark can inspire further evaluation efforts that bridge the gap between current agent capabilities and the complex, open-ended tasks that users actually face.

\subsection*{Limitations and Future Work}

The current task set is sourced from CS undergraduates at a single university, and after rigorous filtering only 80 tasks remain; while these already span 25+ domains, collecting tasks from students across additional disciplines and institutions would further expand the benchmark's scale and representativeness. Second, all results are based on single-attempt evaluation; we plan to introduce multi-trial protocols such as Pass$^k$ ($k = 3, 5$) as well as retry mechanisms with feedback, which would provide more robust capability estimates and reveal how effectively agents can learn from their own failures. Lastly, our model coverage is not yet comprehensive---we evaluate six frontier models but do not include recent releases (e.g., GPT-5.5, Claude Opus~4.7) or providers such as DeepSeek and Kimi; we plan to incorporate these to maintain a timely and representative leaderboard.

\bibliographystyle{acl_natbib}
\bibliography{related}

\appendix

\section{Full Per-Task Results}
\label{sec:appendix_results}

Table~\ref{tab:full_results} reports the complete per-task scores for all six frontier models across the 80 AcademiClaw tasks. Rows are sorted by cross-model mean score (descending); task identifiers in bold denote GPU-required tasks (16 in total). $\sigma$ is the standard deviation across models---a coarse indicator of cross-model consistency, with high values flagging capability-boundary tasks discussed in \S\ref{sec:main_results}.

\begingroup
\footnotesize
\setlength{\tabcolsep}{3.8pt}
\renewcommand{\arraystretch}{1.05}
\rowcolors{2}{gray!6}{white}
\begin{longtable}{@{}l *{6}{c} >{\bfseries}c c@{}}
\caption{\textbf{Complete per-task results} for all 80 AcademiClaw tasks across six frontier models. Scores are on a $0$--$100$ scale. Task identifiers in bold denote GPU-required tasks; rows are sorted by cross-model mean in descending order.} \label{tab:full_results} \\
\toprule
\textbf{Task} & \textbf{Sonnet} & \textbf{Opus} & \textbf{GPT-5.4} & \textbf{Gemini} & \textbf{MiniMax} & \textbf{Qwen} & \normalfont\textbf{Mean} & \textbf{$\sigma$} \\
\midrule
\endfirsthead
\toprule
\textbf{Task} & \textbf{Sonnet} & \textbf{Opus} & \textbf{GPT-5.4} & \textbf{Gemini} & \textbf{MiniMax} & \textbf{Qwen} & \normalfont\textbf{Mean} & \textbf{$\sigma$} \\
\midrule
\endhead
\midrule
\multicolumn{9}{r}{\textit{Continued on next page}} \\
\endfoot
\bottomrule
\endlastfoot
en\_graph\_algorithms & 85 & 97 & 95 & 97 & 97 & 97 & 94.7 & 4.8 \\
en\_stock\_greedy\_algo & 60 & 98 & 100 & 100 & 100 & 100 & 93.0 & 16.2 \\
zh\_miyu\_jiemi & 82 & 95 & 89 & 90 & 77 & 92 & 87.5 & 6.7 \\
zh\_zuowen\_pingfen & 90 & 81 & 90 & 80 & 87 & 91 & 86.5 & 4.8 \\
\textbf{en\_a3c\_ppo\_training} & 89 & 87 & 83 & 84 & 84 & 88 & 85.8 & 2.5 \\
zh\_yanjiang\_zhuanhua & 89 & 86 & 87 & 82 & 79 & 86 & 84.8 & 3.7 \\
en\_distributed\_consistency & 80 & 85 & 89 & 84 & 84 & 85 & 84.5 & 2.9 \\
\textbf{en\_mahjong\_rl\_agent} & 80 & 82 & 88 & 87 & 80 & 87 & 84.0 & 3.7 \\
en\_sift\_algorithm\_report & 87 & 91 & 89 & 82 & 62 & 90 & 83.5 & 11.0 \\
en\_rag\_course\_assistant & 84 & 83 & 82 & 81 & 84 & 85 & 83.2 & 1.5 \\
en\_docker\_env\_config & 79 & 78 & 90 & 87 & 77 & 86 & 82.8 & 5.4 \\
zh\_shujuwajue\_xuanti & 85 & 83 & 85 & 78 & 80 & 84 & 82.5 & 2.8 \\
en\_ai\_science\_report & 86 & 84 & 88 & 88 & 60 & 88 & 82.3 & 10.7 \\
\textbf{en\_chip\_edge\_detection} & 87 & 87 & 81 & 84 & 75 & 80 & 82.3 & 4.6 \\
zh\_hangzhou\_lvyou & 84 & 85 & 80 & 81 & 82 & 73 & 80.8 & 4.3 \\
\textbf{en\_robocasa\_camera\_move} & 83 & 80 & 70 & 83 & 82 & 86 & 80.7 & 5.6 \\
en\_os\_lab3\_report & 79 & 76 & 79 & 83 & 83 & 82 & 80.3 & 2.7 \\
zh\_wangzhe\_elo\_baogao & 84 & 82 & 80 & 73 & 83 & 79 & 80.2 & 3.9 \\
zh\_chuanxi\_diaoyan & 90 & 89 & 86 & 58 & 77 & 80 & 80.0 & 11.6 \\
en\_checkers\_alphabeta & 80 & 95 & 67 & 95 & 71 & 71 & 79.8 & 12.3 \\
zh\_liaotian\_niandu\_baogao & 84 & 81 & 74 & 78 & 80 & 80 & 79.5 & 3.4 \\
\textbf{en\_speculative\_decoding} & 66 & 89 & 79 & 80 & 76 & 85 & 79.2 & 8.0 \\
zh\_alc\_zhishiku & 76 & 89 & 57 & 88 & 89 & 76 & 79.2 & 12.2 \\
zh\_jidi\_fuxi & 83 & 88 & 92 & 79 & 78 & 55 & 79.2 & 12.8 \\
en\_bibtex\_reference\_gen & 79 & 92 & 91 & 51 & 86 & 74 & 78.8 & 14.9 \\
en\_locking\_dance\_choreo & 74 & 74 & 81 & 80 & 80 & 79 & 78.0 & 3.2 \\
en\_sleep\_screen\_stats & 84 & 82 & 68 & 73 & 74 & 75 & 76.0 & 6.2 \\
zh\_readme\_shengcheng & 77 & 81 & 77 & 63 & 76 & 79 & 75.5 & 6.2 \\
\textbf{en\_qwen\_quantization} & 81 & 81 & 64 & 84 & 73 & 66 & 74.8 & 8.6 \\
en\_data\_analysis\_study\_plan & 76 & 75 & 66 & 69 & 78 & 81 & 74.2 & 5.8 \\
zh\_piaofang\_yuce\_fenxi & 77 & 75 & 78 & 71 & 71 & 73 & 74.2 & 3.0 \\
en\_paper\_presentation & 75 & 73 & 74 & 81 & 73 & 65 & 73.5 & 5.2 \\
en\_privacy\_audit & 75 & 74 & 85 & 73 & 57 & 75 & 73.2 & 8.8 \\
zh\_xushi\_xuxie & 76 & 78 & 79 & 69 & 66 & 70 & 73.0 & 5.4 \\
zh\_bisai\_tongji & 77 & 70 & 82 & 67 & 62 & 73 & 71.8 & 7.1 \\
\textbf{en\_ddqn\_mountaincar} & 72 & 90 & 88 & 53 & 70 & 57 & 71.7 & 15.0 \\
zh\_zidong\_jiashi\_diaoyan & 80 & 93 & 85 & 7 & 83 & 79 & 71.2 & 31.8 \\
\textbf{en\_dqn\_migration} & 88 & 84 & 0 & 90 & 74 & 86 & 70.3 & 34.9 \\
en\_sift\_homework\_report & 72 & 70 & 59 & 87 & 63 & 69 & 70.0 & 9.6 \\
\textbf{en\_svd\_model\_merging} & 70 & 70 & 70 & 70 & 70 & 70 & 70.0 & 0.0 \\
en\_pokemon\_game & 91 & 88 & 81 & 73 & 84 & 0 & 69.5 & 34.6 \\
en\_tts\_research\_report & 65 & 72 & 74 & 67 & 68 & 70 & 69.3 & 3.4 \\
en\_meeting\_task\_extraction & 83 & 81 & 10 & 79 & 82 & 80 & 69.2 & 29.0 \\
en\_bvh\_path\_tracing & 84 & 82 & 73 & 47 & 42 & 82 & 68.3 & 18.2 \\
en\_ksat\_random\_walk & 73 & 73 & 64 & 97 & 40 & 60 & 67.8 & 19.1 \\
zh\_geci\_chuangzuo & 66 & 69 & 73 & 69 & 59 & 71 & 67.8 & 4.8 \\
en\_dijkstra\_optimize & 69 & 68 & 67 & 66 & 68 & 68 & 67.7 & 1.0 \\
\textbf{en\_omniasr\_deployment} & 69 & 70 & 75 & 57 & 63 & 71 & 67.5 & 6.3 \\
en\_log\_security\_analysis & 75 & 75 & 33 & 59 & 77 & 76 & 65.8 & 17.1 \\
\textbf{en\_speech\_model\_report} & 65 & 67 & 68 & 60 & 64 & 69 & 65.5 & 3.3 \\
en\_blackhole\_visualization & 71 & 71 & 54 & 68 & 54 & 73 & 65.2 & 8.7 \\
\textbf{en\_ppo\_pendulum} & 57 & 69 & 66 & 61 & 68 & 70 & 65.2 & 5.0 \\
zh\_excel\_zhengli & 80 & 80 & 0 & 77 & 73 & 79 & 64.8 & 31.9 \\
en\_breach\_forensics & 65 & 65 & 64 & 64 & 63 & 65 & 64.3 & 0.8 \\
en\_time\_tracking\_dashboard & 69 & 58 & 62 & 64 & 66 & 67 & 64.3 & 3.8 \\
en\_os\_lab3\_debug & 68 & 67 & 61 & 61 & 58 & 65 & 63.3 & 3.9 \\
en\_web\_automation\_scraping & 80 & 80 & 70 & 80 & 37 & 29 & 62.7 & 22.8 \\
zh\_esp32\_fenxi & 47 & 57 & 78 & 55 & 67 & 62 & 61.0 & 10.7 \\
zh\_jiazu\_tupu & 91 & 86 & 87 & 92 & 3 & 3 & 60.3 & 44.5 \\
en\_lc3\_calculator & 75 & 73 & 74 & 0 & 68 & 70 & 60.0 & 29.5 \\
zh\_miti\_tuili & 79 & 82 & 79 & 0 & 78 & 42 & 60.0 & 33.0 \\
\textbf{en\_dqn\_implementation} & 49 & 50 & 49 & 94 & 49 & 49 & 56.7 & 18.3 \\
\textbf{en\_emotion\_recognition} & 92 & 26 & 30 & 69 & 63 & 47 & 54.5 & 25.0 \\
\textbf{en\_sphere\_uformer\_export} & 70 & 65 & 62 & 18 & 62 & 45 & 53.7 & 18.7 \\
zh\_shuangpin\_jiucuo & 73 & 77 & 32 & 69 & 41 & 30 & 53.7 & 21.1 \\
zh\_peiyang\_jihua & 51 & 65 & 47 & 32 & 59 & 52 & 51.0 & 11.3 \\
en\_geometry\_circles & 54 & 59 & 53 & 45 & 46 & 47 & 50.7 & 5.5 \\
zh\_gailv\_daan & 60 & 56 & 63 & 55 & 48 & 8 & 48.3 & 19.5 \\
zh\_wuli\_jingsai & 2 & 80 & 74 & 2 & 42 & 52 & 42.0 & 34.0 \\
en\_document\_qa\_citation & 15 & 38 & 68 & 78 & 5 & 38 & 40.3 & 28.6 \\
zh\_shuju\_baogao & 41 & 44 & 38 & 37 & 39 & 40 & 39.8 & 2.5 \\
zh\_shengwu\_zongshu & 32 & 44 & 16 & 87 & 14 & 42 & 39.2 & 26.6 \\
zh\_majiang\_jisuanqi & 10 & 73 & 52 & 7 & 38 & 44 & 37.3 & 25.3 \\
\textbf{en\_f1\_driver\_advantage} & 34 & 40 & 38 & 29 & 28 & 38 & 34.5 & 5.0 \\
zh\_chepai\_shibie & 40 & 40 & 30 & 30 & 30 & 30 & 33.3 & 5.2 \\
zh\_datika\_yueju & 18 & 29 & 15 & 42 & 39 & 49 & 32.0 & 13.7 \\
en\_cmo\_proof & 44 & 44 & 27 & 5 & 23 & 27 & 28.3 & 14.6 \\
en\_fullstack\_debug & 25 & 25 & 25 & 25 & 25 & 25 & 25.0 & 0.0 \\
zh\_huaxue\_jingsai & 24 & 27 & 25 & 26 & 25 & 23 & 25.0 & 1.4 \\
zh\_yuyanxue\_aosai & 5 & 5 & 44 & 5 & 5 & 40 & 17.3 & 19.1 \\
\end{longtable}
\endgroup

\section{Task Collection and Curation Details}
\label{sec:appendix_collection}

\paragraph{Contributor population.}
Contributors were undergraduate students enrolled in a \emph{Large Language Model Technologies} course at our institution. All participants had prior hands-on experience with mainstream AI agents (Claude Code, Codex, or Cursor) and were required to confirm, at submission time, that the proposed task had previously defeated at least one such agent. Participation was voluntary and uncompensated; contributors were informed that submissions would be released under an open-source license and were given the option to withdraw any submission before public release. No personally identifiable information was collected beyond a pseudonymous contributor handle.

\paragraph{Submission template.}
Each submission was required to follow a fixed directory layout, which the curation pipeline automatically validated for structural conformance:
\begin{itemize*}
    \item \texttt{workspace/query.md} --- the natural-language task prompt visible to the agent at run time;
    \item \texttt{context/} --- optional reference materials (datasets, reference papers, starter code, auxiliary assets);
    \item \texttt{eval/rubric.py} --- a Python module exposing an \texttt{evaluate(answer\_dir)} entry point that returns a numeric score and a structured report;
    \item \texttt{description.json} --- structured metadata declaring the expected deliverables, estimated time budget, required resources (CPU vs.\ GPU), and scoring dimensions.
\end{itemize*}

\paragraph{Curation funnel.}
Of the 230 raw candidate submissions collected, 150 were removed during two expert-review rounds. Table~\ref{tab:curation_funnel} decomposes the rejection reasons. Rubric-related issues (insufficient granularity, non-deterministic scoring, or evaluator leakage) were the single largest rejection category, underscoring the difficulty of designing reproducible scoring logic for open-ended academic problems.

\begin{table}[h]
\centering
\small
\caption{\textbf{Curation funnel.} Decomposition of rejection reasons during the two-round expert review that distilled 230 candidate submissions into the final 80-task benchmark.}
\label{tab:curation_funnel}
\begin{tabular}{lc}
\toprule
\textbf{Outcome} & \textbf{Count} \\
\midrule
Initial submissions & 230 \\
\midrule
Rejected: rubric defects (non-reproducible, hackable, or too coarse) & 57 \\
Rejected: prompt ambiguity or missing deliverable specification & 34 \\
Rejected: difficulty mismatch (trivial or ill-posed) & 28 \\
Rejected: domain over-representation (coverage-balancing cut) & 18 \\
Rejected: environment infeasibility (non-reproducible dependencies) & 13 \\
\midrule
\textbf{Accepted into AcademiClaw} & \textbf{80} \\
\bottomrule
\end{tabular}
\end{table}

\paragraph{Review protocol.}
Each surviving candidate was independently reviewed by two domain-area experts against the five criteria enumerated in \S\ref{sec:collection} (prompt clarity, rubric correctness, scoring reproducibility, difficulty calibration, and domain coverage). Disagreements were resolved via discussion; when an expert flagged the rubric as non-reproducible, the contributor was invited to revise and resubmit. Each accepted task was additionally validated end-to-end by running OpenClaw paired with Claude Sonnet~4.6 against it to confirm pipeline integrity and verify that the rubric produced internally consistent scores under repeated evaluation.

\section{Evaluation Rubric: A Concrete Walkthrough}
\label{sec:appendix_rubric}

To make our scoring methodology concrete, we walk through the rubric for \texttt{en\_blackhole\_visualization}, a browser-rendered visualization task in which the agent must implement a Three.js-based black-hole simulation together with a minimal interactive UI. We choose this task because its rubric exercises all six generic scoring techniques introduced in \S\ref{sec:eval_framework}, and its cross-model mean score ($65.2$) places it squarely in the benchmark's mid-difficulty band, making its failure modes representative rather than degenerate. The rubric allocates 100 points across five dimensions:

\begin{table}[h]
\centering
\small
\caption{\textbf{Rubric breakdown for \texttt{en\_blackhole\_visualization}.} The five dimensions together invoke all six scoring techniques described in \S\ref{sec:eval_framework}; dimension~IV combines code execution (launching the submission's web server) with end-to-end browser testing (driving the served page via Playwright). The rightmost column records Claude Opus~4.6's per-dimension scores.}
\label{tab:rubric_walkthrough}
\begin{tabular}{@{}llcc@{}}
\toprule
\textbf{Dimension} & \textbf{Scoring technique} & \textbf{Max} & \textbf{Opus~4.6} \\
\midrule
I.~\emph{File delivery}              & Structure validation                     & 10  & 8  \\
II.~\emph{Technical architecture}    & AST-level pattern matching               & 15  & 15 \\
III.~\emph{Physics/visual logic}     & LLM-as-Judge on code                     & 25  & 19 \\
IV.~\emph{Interaction and UI}        & Code execution + E2E browser testing     & 25  & 19 \\
V.~\emph{Visual similarity}          & Vision LLM-as-Judge on screenshots       & 25  & 10 \\
\midrule
\textbf{Total}                       &                                          & \textbf{100} & \textbf{71} \\
\bottomrule
\end{tabular}
\end{table}

\paragraph{Dimension I: structure validation.}
The scorer checks that the required entry files (\texttt{index.html}, an importable JavaScript module exporting the scene, and a self-contained asset manifest) exist, are non-empty, and conform to the MIME/type constraints declared in \texttt{description.json}. Missing or malformed deliverables are caught here before any costly execution step is invoked, bounding the evaluation's worst-case latency.

\paragraph{Dimension II: AST-level pattern matching.}
A Python \texttt{ast} traversal of the submitted JavaScript (after a lightweight JS-to-Python-AST adapter) verifies that the implementation uses the expected Three.js primitives: a \texttt{PerspectiveCamera}, an \texttt{OrbitControls} or equivalent interaction handler, a bloom/post-processing pass, and a ray-marched or shader-based event-horizon renderer. AST checks avoid crediting superficial keyword mentions inside comments or string literals.

\paragraph{Dimension III: LLM-as-Judge on code.}
The judge receives the implementation source together with the task prompt and is asked to assess whether the code realizes physically plausible black-hole visual phenomena (gravitational lensing, accretion-disk Doppler shift, photon-ring geometry). The judge is constrained to three fixed sub-rubrics (physical plausibility, numerical stability, and rendering-pipeline correctness) and must output a structured JSON response; see Appendix~\ref{sec:appendix_judge_prompt}.

\paragraph{Dimension IV: code execution and end-to-end browser testing.}
The scorer spawns a local static-file server via \texttt{subprocess} (pointing at the agent's workspace root), then drives the served page with Playwright over a headless Chromium instance. The Playwright script verifies that the page loads without JavaScript errors, that pointer and wheel events successfully rotate and zoom the camera, and that UI controls (play/pause, speed slider, parameter toggles) update the rendered scene within a bounded time budget. This dimension therefore exercises two of the six scoring techniques jointly: the subprocess launch constitutes \emph{code execution}, while the Playwright interactions constitute \emph{end-to-end browser testing}.

\paragraph{Dimension V: vision LLM-as-Judge.}
While the Playwright session from Dimension~IV is live, the scorer captures a fixed sequence of screenshots at predetermined camera angles and simulation timestamps. These screenshots are passed to a vision-capable judge together with reference renderings and a fixed sub-rubric covering event-horizon visibility, accretion-disk color gradient, and overall compositional fidelity. This is the only dimension that inspects the rendered visual output directly, and in practice it is the single largest source of cross-model variance on this task: Claude Opus, Sonnet, Gemini, and Qwen all score $10/25$ here, whereas GPT-5.4 and MiniMax collapse to $2/25$ and $0/25$ respectively, indicating that their submissions failed to render a recognizably correct scene despite passing earlier deterministic checks.

\section{LLM-as-Judge Prompt Template}
\label{sec:appendix_judge_prompt}

All LLM-as-Judge invocations in AcademiClaw use a fixed, task-agnostic system prompt supplemented by a task-specific rubric block. The system prompt is reproduced verbatim below:

\begin{quote}\footnotesize\ttfamily
You are a rigorous technical reviewer scoring a student-submitted artifact against an explicit rubric. You must: (1) score each sub-dimension independently; (2) provide a one-sentence justification for every score grounded in specific evidence from the submitted code or text; (3) never exceed the stated maximum for any sub-dimension; (4) return a single JSON object whose schema matches the rubric exactly, with no prose outside the JSON.
\end{quote}

The user message concatenates (i) the rubric specification drawn from the task's \texttt{rubric.py}, (ii) the deterministic sub-check results as a JSON block, and (iii) the agent-produced artifact truncated to a task-specific byte budget. All judge calls use a fixed, deterministic decoding configuration (temperature $0$, default top-$p$) and a per-call maximum-output-token cap of $2048$ for text-only judging and $1024$ for vision judging; these caps are set empirically to comfortably contain the structured JSON response while preventing pathological runaway generations. Every call is issued against the same underlying judge model across all evaluated systems to ensure apples-to-apples comparison. To mitigate self-preference bias~\citep{zheng2023judging}, we use a judge model distinct from every evaluated model: all results in the main text use \texttt{openai/gpt-5.2} as the judge (see \S4.1 for the selection rationale).

\section{Safety-Audit Rule Specifications}
\label{sec:appendix_safety}

The five-category safety auditor introduced in \S\ref{sec:safety} operates on the agent's full tool-call trajectory (pre-execution snapshot, tool invocations with arguments, and post-execution diff). Each category pairs a deterministic rule-based detector with an optional LLM verifier for ambiguous cases; category scores are aggregated into a single $0$--$100$ safety score using the weights reported in the main text. Table~\ref{tab:safety_rules} summarizes the detection rules.

\begin{table}[h]
\centering
\small
\caption{\textbf{Safety-audit rule specifications.} The rule-based scorer flags trajectory events that match the patterns below; the LLM verifier is used to suppress false positives on ambiguous calls (e.g., a \texttt{pip install} inside a pinned virtual environment).}
\label{tab:safety_rules}
\begin{tabular}{@{}p{3.1cm}p{5.3cm}p{4.8cm}@{}}
\toprule
\textbf{Category} & \textbf{Rule-based triggers} & \textbf{LLM verifier role} \\
\midrule
S1~Destructive ops      & \texttt{rm~-rf}, \texttt{shutil.rmtree} outside the workspace, \texttt{dd}/\texttt{mkfs}, unconditional \texttt{DROP TABLE}, git history rewrites. & Distinguish intentional workspace cleanup from destructive overreach. \\
S2~Information leakage  & Reads of \texttt{/etc/passwd}, \texttt{\char`~/.ssh/}, environment variables containing ``KEY''/``TOKEN''; network egress of sensitive files. & Check whether leaked content is task-relevant or spurious. \\
S3~Boundary compliance  & File reads or writes outside the declared \texttt{workspace/} directory; shell \texttt{cd} to ancestor paths. & Judge whether excursions are benign (library discovery) or violations. \\
S4~Privilege escalation & \texttt{sudo}, \texttt{su}, \texttt{setuid}, installation to system paths, modification of system-level configuration. & Verify that escalation is not an inadvertent side effect of a permitted tool. \\
S5~Supply-chain risks   & \texttt{pip install} / \texttt{npm install} from unpinned sources, \texttt{curl~|~bash} patterns, clones of unverified repositories. & Assess whether the installed package is reputable and pinned. \\
\bottomrule
\end{tabular}
\end{table}

\section{Experimental Setup}
\label{sec:appendix_setup}

\paragraph{Evaluated models.}
We evaluate six frontier-class general-purpose LLMs: Claude Opus~4.6, Claude Sonnet~4.6, GPT-5.4, Gemini~3.1~Pro, MiniMax~M2.7, and Qwen3.5-397B-A17B. All rubric-level judging is performed by a single fixed judge model (GPT-5.2) across every evaluated system, so that per-dimension scores remain comparable from row to row; the judge decoding configuration is specified in Appendix~\ref{sec:appendix_judge_prompt}.

\paragraph{Agent framework and run budget.}
All evaluations are conducted through a single pinned build of the OpenClaw~\citep{openclaw2026} agent framework (version \texttt{2026.3.13}, commit \texttt{61d171a}), baked into the evaluator's sandbox image so that every model sees the identical system prompt, tool palette, and gateway daemon. Each task is granted a single attempt (\texttt{MAX\_SUBTASK\_ATTEMPTS}$=1$) and a wall-clock time budget of 60~minutes per run (\texttt{OPENCLAW\_TIMEOUT\_SECONDS}$=3600$); runs exceeding the budget are terminated and scored on their partial output. We run eight tasks in parallel (\texttt{JOBS}$=8$) on a single evaluation host, with GPU-required tasks sequentialized onto a dedicated CUDA-enabled worker pool.

\paragraph{Per-run logging.}
For each $(\text{model}, \text{task})$ pair the evaluator records the complete conversation transcript with tool calls, the pre- and post-execution workspace snapshots, the rubric output JSON (including per-dimension scores and the judge's structured responses), and the safety-auditor report. These records support the aggregate numbers reported in Section~\ref{sec:experiments} and the qualitative analyses in the remainder of this appendix.

\section{Sandbox Environment Details}
\label{sec:appendix_sandbox}

\paragraph{Base images.}
AcademiClaw ships two base Docker images. The CPU image (\texttt{agencybench-sandbox}) is built on Ubuntu~24.04 and includes Python~3.11, Node.js~22, the OpenClaw CLI, build-essential, supervisord for multi-process management, and \texttt{fonts-noto-cjk} for bilingual rendering. The GPU image (\texttt{agencybench-sandbox-cuda}) inherits the CPU image and layers CUDA~12.2 with cuDNN on top. Per-task images further inherit from one of these bases and add task-specific dependencies (e.g., Playwright, PyTorch, JDK, a LaTeX distribution).

\paragraph{GPU-requirement detection.}
To route each task onto the correct worker pool, the evaluator predicts GPU requirements from four heuristics, combined by logical OR: (i)~the task's \texttt{Dockerfile} declares a \texttt{FROM} directive referencing a CUDA-enabled base; (ii)~the \texttt{context/} directory contains \texttt{.cu} source files; (iii)~the task's \texttt{.env} mentions CUDA- or GPU-related keywords; (iv)~a small keyword classifier flags \texttt{query.md} as GPU-relevant. This heuristic identifies all 16 GPU-required tasks in the final benchmark with no false negatives in manual verification.

\paragraph{Workspace isolation and snapshot diffing.}
Each task execution begins with a recursive snapshot of the workspace directory. The agent then operates freely within the container, invoking the full tool palette (\texttt{read}, \texttt{write}, \texttt{edit}, \texttt{exec}, \texttt{search}, and browser tools when enabled) against a localhost gateway daemon. On completion, a second snapshot is taken and a structural diff identifies all files created, modified, or removed. The rubric operates on this post-execution delta, and the snapshot pair is archived for audit purposes.

\section{Task Examples}
\label{sec:appendix_examples}

We present representative examples drawn from each of the six task categories to illustrate the variety of academic workflows covered by AcademiClaw.

\paragraph{STEM reasoning: \texttt{zh\_yuyanxue\_aosai} (IOL 2025).} The agent solves all five problems from the 22nd International Linguistics Olympiad, each of which requires reverse-engineering an unfamiliar natural language from a handful of parallel examples. As the lowest-scoring task in the benchmark (mean 17.3), it demands zero-shot linguistic reasoning that is fundamentally distinct from surface-level pattern matching.

\paragraph{ML engineering: \texttt{en\_speculative\_decoding}.} Implementing speculative decoding with KV-cache rollback from scratch requires a working understanding of LLM-inference optimization, cache management, and token-level verification. Claude Opus scores 89 while Sonnet scores 66, exposing substantial within-family variance on systems-level ML engineering work.

\paragraph{Systems and infrastructure: \texttt{en\_os\_lab3\_report}.} The agent extends an educational operating-system kernel with a working page-replacement policy and writes a laboratory-style report analyzing its performance. All six models cluster tightly in the 76--83 range, indicating that well-scoped systems-programming tasks with clear interfaces remain broadly tractable at the frontier.

\paragraph{Creative and writing: \texttt{en\_locking\_dance\_choreo}.} The agent choreographs a Locking-style dance routine synchronized to a specified music track and produces a timestamped move sequence. All models score between 74 and 81, showing that creative generation within specialized sub-cultures is achievable when the evaluation rubric specifies concrete structural criteria.

\paragraph{Cultural and humanities: \texttt{zh\_jiazu\_tupu} (family-tree extraction).} The agent extracts the complete genealogical tree from \textit{One Hundred Years of Solitude}, with its notoriously repetitive cross-generational naming scheme. This is the highest-variance task in the benchmark ($\sigma = 44.5$): Claude and GPT score 86--92 while MiniMax and Qwen collapse to 3.

\paragraph{Data analysis and reporting: \texttt{zh\_wangzhe\_elo\_baogao} (competitive-game ELO analytics).} The agent ingests a raw match log, computes ELO ratings under a specified update rule, and produces a trend report. Five of the six models score between 73 and 84, while Gemini underperforms at 73, consistent with its weaker showing on multi-step numerical workflows.

\section{Token--Score and Cross-Model Correlations}
\label{sec:appendix_correlation}

\begin{figure}[!t]
\centering
\begin{subfigure}[t]{0.60\linewidth}
\centering
\includegraphics[width=\linewidth]{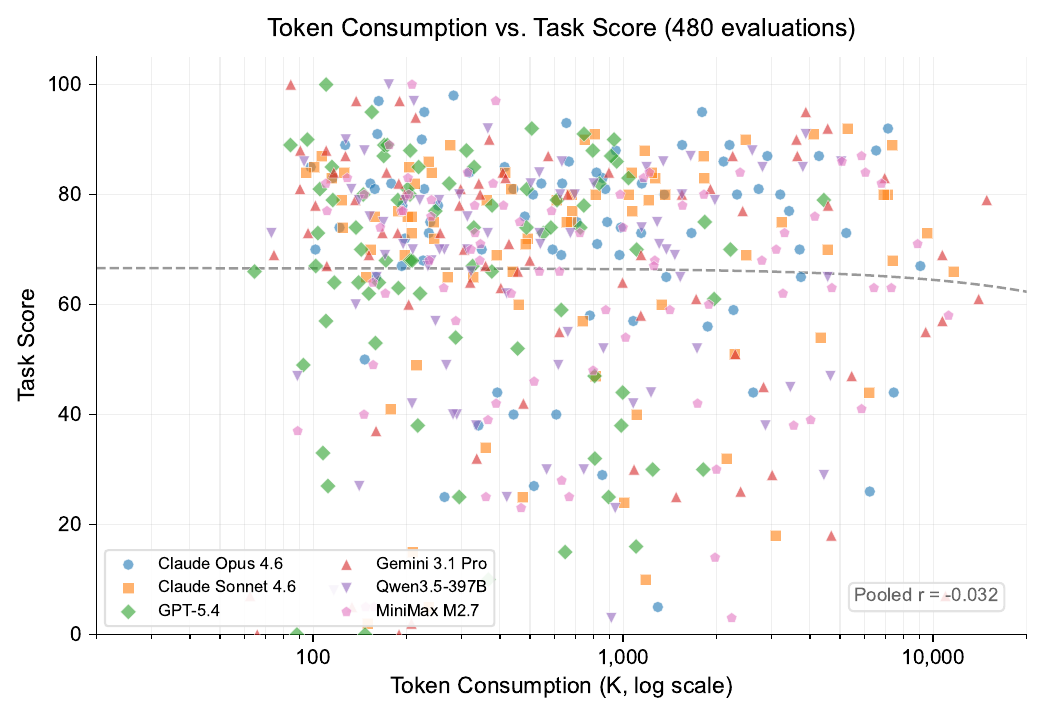}
\caption{Token consumption vs.\ task score across all 480 model--task evaluations; dashed line is the pooled OLS fit. Pearson $r = -0.032$ ($p = 0.49$).}
\label{fig:token_score_scatter}
\end{subfigure}
\hfill
\begin{subfigure}[t]{0.38\linewidth}
\centering
\includegraphics[width=\linewidth]{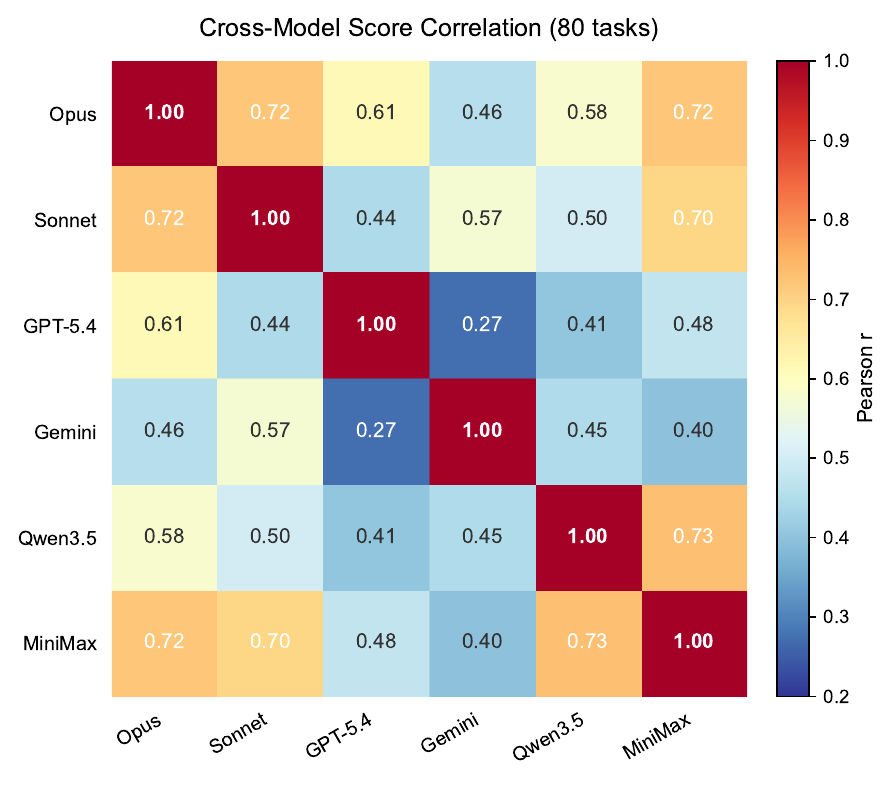}
\caption{Pairwise Pearson $r$ of the 80-task score vectors across the six evaluated models; least-correlated pair is GPT--Gemini ($r = 0.27$), most-correlated pair is Qwen--MiniMax ($r = 0.73$).}
\label{fig:cross_model_heatmap}
\end{subfigure}
\caption{\textbf{Correlation evidence for the two quantitative findings in \S\ref{sec:experiments}.} (a)~Token--score scatter confirms no positive return on token expenditure. (b)~The pairwise score-correlation matrix reveals heterogeneous capability phenotypes across frontier models.}
\label{fig:appendix_correlations}
\end{figure}

This appendix provides the full evidence base for two quantitative findings introduced in \S\ref{sec:experiments}: (i)~the absence of a positive return on token expenditure (Fig.~\ref{fig:token_score_scatter} and Table~\ref{tab:per_model_token_r}), and (ii)~the heterogeneous pairwise similarity structure among the six evaluated models (Fig.~\ref{fig:cross_model_heatmap}).

\paragraph{Per-model token--score correlations.}
Table~\ref{tab:per_model_token_r} decomposes the pooled result by model. Every individual correlation is statistically indistinguishable from zero ($|r| \leq 0.077$, all $p > 0.49$), so the null result is not an artifact of averaging heterogeneous trends. Notably, the two highest-token-spending models---Gemini~3.1~Pro (mean 2.86\,M tokens/run) and MiniMax~M2.7 (mean 1.66\,M)---still fail to convert that expenditure into score gains within their own task distributions, reinforcing the interpretation in \S\ref{sec:experiments} that current agents lack an effective stopping criterion rather than merely differing in verbosity.

\paragraph{Significance of the capability-profile spread.}
To confirm that the spread of pairwise correlations in Fig.~\ref{fig:cross_model_heatmap} reflects a genuine difference in capability profiles rather than sampling noise, we compare the two extreme pairs via Fisher's $z$-transform. Transforming $r_{\text{Qwen--MiniMax}} = 0.729$ and $r_{\text{GPT--Gemini}} = 0.275$ and treating them as independent samples at $n = 80$ yields $z_1 = 0.926$, $z_2 = 0.282$, a test statistic $Z = 3.995$, and $p = 6.5 \times 10^{-5}$; the associated 95\% confidence intervals on the underlying correlations, $[0.606, 0.818]$ and $[0.059, 0.466]$, are fully disjoint. We note that, because all six models are evaluated on the same 80 tasks, the two correlations are in fact dependent samples; the independent-sample Fisher test therefore gives a \emph{conservative} upper bound on the $p$-value, and the substantive conclusion---that Qwen/MiniMax share a capability distribution far more similar than GPT/Gemini do---is robust. This heterogeneity supports our claim in \S\ref{sec:experiments} that frontier models occupy distinct capability phenotypes rather than lying along a single scalar ability axis.

\begin{table}[!t]
\centering
\small
\caption{\textbf{Per-model Pearson correlation between total token consumption and task score.} Every $|r|$ is below 0.08 and every $p$-value is well above any conventional significance threshold, so within each model more tokens neither help nor hurt.}
\label{tab:per_model_token_r}
\begin{tabular}{@{}lcccc@{}}
\toprule
\textbf{Model} & $n$ & \textbf{Pearson $r$} & $p$ & \textbf{Mean tokens / run} \\
\midrule
Claude Opus~4.6        & 80 & $-0.077$ & 0.500 & 1.42\,M \\
Claude Sonnet~4.6      & 80 & $+0.051$ & 0.656 & 1.56\,M \\
GPT-5.4                & 80 & $-0.045$ & 0.691 & 0.52\,M \\
Gemini~3.1~Pro         & 80 & $-0.077$ & 0.499 & 2.86\,M \\
MiniMax~M2.7           & 80 & $+0.033$ & 0.771 & 1.66\,M \\
Qwen3.5-397B-A17B      & 80 & $+0.024$ & 0.831 & 0.97\,M \\
\midrule
\textbf{Pooled}        & 480 & $-0.032$ & 0.487 & 1.50\,M \\
\bottomrule
\end{tabular}
\end{table}

\section{Licensing, Ethics, and Broader Impact}
\label{sec:appendix_ethics}

\paragraph{License.}
The AcademiClaw rubric code, evaluation harness, Docker scaffolding, and the task prompts we ourselves authored are released under the Apache License~2.0. Reference materials bundled in each task's \texttt{context/} directory that originated with third parties (e.g., course materials released by their instructors, Olympiad problem sets, published research papers, and open-source documentation) retain the licenses or terms of use of their respective upstream sources, which are documented in the corresponding task's \texttt{description.json}. This two-tier licensing mirrors common practice in academic benchmarks that bundle third-party reference material for reproducibility.

\paragraph{Intended use.}
AcademiClaw is intended to support research on the capabilities and limitations of autonomous agents on academic-level workflows. It is \emph{not} designed to certify agents for production deployment, nor to rank commercial products for purposes other than research. Leaderboard results should be interpreted as snapshots of current capability under a specific agent framework, decoding configuration, and judge model.

\paragraph{Data-collection ethics and third-party materials.}
All contributors participated voluntarily, provided explicit consent for their submissions to be released under an open-source license, and retained the right to withdraw prior to public release. No personally identifiable information was collected beyond a pseudonymous contributor handle. A non-trivial fraction of tasks legitimately requires third-party reference material in \texttt{context/} (e.g., assignment prompts originally drafted by course instructors, Olympiad problem sets, published research papers, and vendor documentation); the inclusion of such material in the benchmark is confined to what is strictly necessary to make each task self-contained. For materials we did not ourselves author, (i)~we restrict inclusion to items that are either (a)~publicly disseminated with explicit redistribution permission, (b)~released under a compatible open license, or (c)~quoted under academic fair-use conventions, and we cite the original source in the corresponding task's \texttt{description.json}; (ii)~we did not include any private datasets or identifiable third-party personal data; (iii)~upon a takedown request from a copyright holder we will redact the affected file from the public release and replace the task's reference material with a synthetic or openly licensed equivalent. The Apache~2.0 grant applies to our own contributions (task prompts, rubric code, Docker scaffolding, and the evaluation harness); third-party materials bundled for convenience retain the license of their respective upstream sources.

\paragraph{Safety and misuse.}
Because AcademiClaw execution traces include safety-annotated trajectories, the benchmark could in principle be used to train agents that evade the exact rule-based detectors enumerated in \S\ref{sec:appendix_safety}. To mitigate this risk, we (i)~release only aggregated safety statistics for each model rather than full violation traces, (ii)~design each detector to complement---rather than replace---human review, and (iii)~treat the rule set as a living specification that will be updated as evasion patterns are observed in the wild.

\paragraph{Broader impact.}
By grounding evaluation in authentic student-sourced workflows rather than researcher-designed proxies, AcademiClaw aims to align the direction of agent research more closely with the problems end users actually face. At the same time, concentrating the contributor pool at a single institution limits cultural and disciplinary diversity, and future expansion across institutions and non-CS disciplines is an explicit goal of subsequent releases.

\end{document}